\documentclass[10pt,journal,compsoc]{IEEEtran}
\ifCLASSOPTIONcompsoc
  \usepackage[nocompress]{cite}
\else
  \usepackage{cite}
\fi
\ifCLASSINFOpdf
\else
\fi
\usepackage{algorithmic}
\usepackage[utf8]{inputenc} 
\usepackage[T1]{fontenc}    
\usepackage{url}            
\usepackage{booktabs}       
\usepackage{amsfonts}       
\usepackage{nicefrac}       
\usepackage{microtype}      
\usepackage{listings}
\usepackage[normalem]{ulem}
\useunder{\uline}{\ul}{}
\usepackage{diagbox}
\usepackage{colortbl}
\usepackage{yh_style}

\definecolor{forestgreen}{rgb}{0.133, 0.545, 0.133}
\definecolor{yellowyellow}{rgb}{0.133, 0.545, 0.133}
\definecolor{ntu_red}{HTML}{7C223F}
\definecolor{ntu_blue}{HTML}{181C62}
\definecolor{COLOR_MEAN}{HTML}{f0f0f0}
\definecolor{ROW_COLOR}{HTML}{C9F7F4}
\definecolor{greendot}{HTML}{06d6a0}
\definecolor{magneta}{HTML}{FE6D73}
\definecolor{dark_green}{HTML}{17C3B2}
\definecolor{wrong_color}{HTML}{FFB29B}
\definecolor{decoding_color}{HTML}{D26379}

\usepackage{marvosym}
\makeatletter
\def\@fnsymbol#1{\ensuremath{\ifcase#1\or \textsuperscript{~\Letter}\or \ddagger\or
   \mathsection\or \mathparagraph\or \|\or **\or \dagger\dagger
   \or \ddagger\ddagger \else\@ctrerr\fi}}
\makeatother

\usepackage[pagebackref,breaklinks,colorlinks,citecolor=citecolor]{hyperref}
\usepackage{threeparttable}
\usepackage{wrapfig}
\usepackage[capitalize]{cleveref}
\crefname{section}{Sec.}{Secs.}
\Crefname{section}{Section}{Sections}
\Crefname{table}{Table}{Tables}
\crefname{table}{Tab.}{Tabs.}

\lstdefinestyle{instructformat}{
  basicstyle=\ttfamily\small,
  frame=single,
  breaklines=true,
  breakindent=0pt, 
  keywordstyle=\color{blue},
  keywords={User, GPT},
  xleftmargin=0pt, 
}

\lstdefinestyle{qaformat}{
  basicstyle=\ttfamily\footnotesize,
  frame=none,
  breaklines=true,
  breakindent=0pt, 
  keywordstyle=\color{blue},
  keywords={Q, A, Prompt, Question, Answer, GT, Response, Score, Short, Long},
  xleftmargin=0pt, 
}

\begin{document}
\title{Otter: A Multi-Modal Model with In-Context Instruction Tuning}
%
%
%
%

\author{Bo Li$^{\ast}$, Yuanhan Zhang$^{\ast}$, Liangyu Chen$^{\ast}$, Jinghao Wang$^{\ast}$, Fanyi Pu$^{\ast}$ \protect \\ Joshua Adrian Cahyono, Jingkang Yang, Chunyuan Li, Ziwei Liu~\Letter
\IEEEcompsocitemizethanks{\IEEEcompsocthanksitem $^{\ast}$Equal Contribution. \protect
\IEEEcompsocthanksitem Bo Li, Yuanhan Zhang, Liangyu Chen, Jinghao Wang, Fanyi Pu, Joshua Adrian Cahyono, Jingkang Yang and Ziwei Liu are with the S-Lab, Nanyang Technological University. \protect \\
E-mail: \{libo0013, yuanhan002, liangyu.chen, jinghao003, fpu001, jo0001no, jingkang001, ziwei.liu\}@ntu.edu.sg
\IEEEcompsocthanksitem  Chunyuan Li is with the Microsoft Research, Redmond. \protect \\
E-mail: chunyuan.li@microsoft.com}}


%
%

\markboth{Journal of \LaTeX\ Class Files,~Vol.~14, No.~8, August~2015}%
{Shell \MakeLowercase{\textit{et al.}}: Bare Demo of IEEEtran.cls for Computer Society Journals}
%



\IEEEtitleabstractindextext{%
\begin{abstract}
Recent advances in Large Multimodal Models (LMMs) have unveiled great potential as visual assistants. However, most existing works focus on responding to individual instructions or using previous dialogues for contextual understanding. There is little discussion on employing both images and text as in-context examples to enhance the instruction following capability.
To bridge this gap, we introduce the \textbf{Otter} model to leverage both textual and visual in-context examples for instruction tuning. 
Specifically, Otter builds upon Flamingo with Perceiver architecture, and has been instruction tuned for general purpose multi-modal assistant. Otter seamlessly processes multi-modal inputs, supporting modalities including text, multiple images, and dynamic video content.
To support the training of Otter, we present the \textbf{MIMIC-IT} (\textbf{M}ult\textbf{I}-\textbf{M}odal \textbf{I}n-\textbf{C}ontext \textbf{I}nstruction \textbf{T}uning) dataset, which encompasses over 3 million multi-modal instruction-response pairs, including approximately 2.2 million unique instructions across a broad spectrum of images and videos. MIMIC-IT has been carefully curated to feature a diverse array of in-context examples for each entry. 
Comprehensive evaluations suggest that instruction tuning with these in-context examples substantially enhances model convergence and generalization capabilities. Notably, the extensive scenario coverage provided by the MIMIC-IT dataset empowers the Otter model to excel in tasks involving complex video and multi-image understanding.
\end{abstract}

}

\maketitle
\IEEEdisplaynontitleabstractindextext

%
\IEEEpeerreviewmaketitle

\section{Introduction}

\begin{figure*}[htp]
    \centering
    \includegraphics[width=\textwidth]{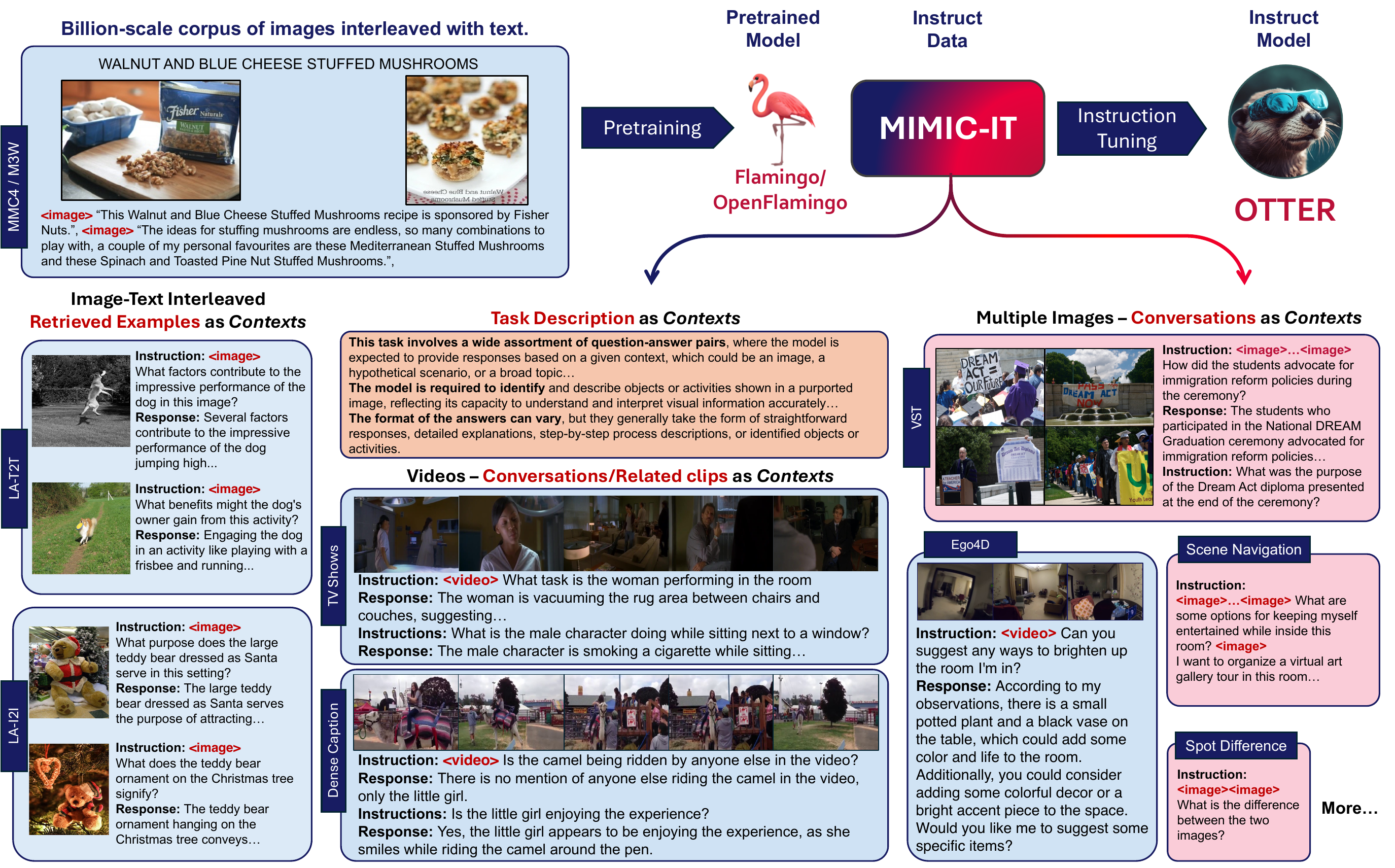}
    \caption{\textcolor{black}{\textbf{Otter}, built upon the Flamingo architecture, leverages in-context instruction tuning to align with Flamingo’s upstream image-text interleaved pretraining paradigm. The \textbf{MIMIC-IT} dataset consists of 3 million multimodal instruction-response pairs, where each instruction is paired with multimodal contexts. This enables LMMs trained on MIMIC-IT to exhibit strong capabilities as general-purpose assistants. The specific data format of MIMIC-IT is shown in~\cref{fig:instrct_template}.}}
    \label{fig:teaser}
\end{figure*}

Recent advancements in Large Language Models (LLMs), exemplified by GPT-2~\cite{gpt2} and GPT-3~\cite{gpt3}, have highlighted their effectiveness as few/zero-shot learners trained on vast text corpora. The development of models like InstructionGPT~\cite{instruct_gpt} and ChatGPT~\cite{gpt4} through instruction tuning has significantly enhanced their capacity for understanding and executing natural language instructions. This innovation integrates task-specific rules during fine-tuning, markedly improving user intent comprehension and response precision. Concurrently, there has been significant progress in multi-modal models~\cite{liu2023prismer,chen2023pali3,chen2023pali,llama_adapater,llava,instruct_blip,chen2023shikra,alayrac2022flamingo,awadalla2023openflamingo,Qwen_VL,mini_gpt4,laurenccon2023obelics,li2023otterhd}, reflecting a trend towards leveraging diverse modalities like images and text. This shift aims to provide a more comprehensive approach to understanding and engaging with the real world.

These evolutions are primarily attributed to the exploration of in instruction tuning~\cite{flan,flant5,wang2022super,self_instruct,alpaca,vicuna2023,peng2023instruction} applied to both Large Language Models (LLMs) and Large Multimodal Models (LMMs). Instruction tuning, a process that refines models through diverse, high-quality instructions~\cite{flant5,self_instruct}, has notably enhanced zero-shot capabilities in natural language processing~\cite{flan} and multimodal tasks~\cite{llama_adapater,llama_adapter_v2,mini_gpt4,li2023blip,dai2023instructblip,liu2023visual}.
However, current instruction tuning concentrates on simple instruction towards single image or leveraging preceding dialogues as contextual information. They may overlook the potential synergistic effect of utilizing both images and text as contextual information during the instruction tuning process. 

Reflecting on multi-modal models like Flamingo, which effectively integrate image-text data, their training underscores the synergy between images and texts. This interplay motivates our adoption of a similar approach in instruction tuning, combining both modalities to in-context examples.
Furthermore, studies emphasize the benefits of optimized contextual information in contrastively learned visual language models~\cite{zhou2022cocoop,zhou2022coop}. Meanwhile, in language models, instruction tuning datasets such as Flan collection~\cite{flan} offer rich, complex textual contexts per instance (\textit{e.g.} \textit{Premise}, \textit{Hypothesis}, and \textit{Target}), this approach would potentially improve instruction understanding and facilitate differentiation from other instructions or tasks. 

Inspired by these insights, we introduce the \textbf{Otter} model, inherited from the Flamingo structure, designed to effectively leverage this multimodal synergy and instruction-tuned to function as a general-purpose multimodal assistant. The in-context instruction tuning on Otter aligns seamlessly with Flamingo’s upstream image-text interleaved pretraining paradigm. This approach enhances the model's proficiency in comprehending and executing complex instructions and improving training efficacy.

To better facilitate the in-context instruction tuning of Otter model, we propose \textbf{MIMIC-IT} (\textbf{M}ult\textbf{I}-\textbf{M}odal \textbf{I}n-\textbf{C}ontext \textbf{I}nstruction \textbf{T}uning) dataset. This expansive dataset encompasses over 3 million multimodal instruction-response pairs, including approximately 2.2 million unique instructions that span a wide spectrum of images and videos. The MIMIC-IT dataset stands out for its thorough curation, featuring a diverse array of multi-modal contextual information for each instance. MIMIC-IT's broad and diverse scenario coverage enables Otter to excel in complex tasks involving video and multi-image comprehension. The dataset's rich contextual variety forms the backbone of the model's proficiency in these domains, highlighting its adaptability and wide-ranging applicability. 

Also as evidenced in~\cref{sec:experiments}, in-context instruction tuning significantly boosts Otter's training convergence and generalization capabilities. This is reflected in the our analysis and Otter's superior performance across various benchmarks. We concretize our contributinos as follows:

\begin{itemize}
    \setlength{\itemsep}{4pt}
    \setlength{\parsep}{0pt}
    \setlength{\parskip}{0pt}
    \item \textbf{Otter Model:} We introduce the \textbf{Otter}, a multi-modality model crafted to leverage both textual and visual in-context examples during instruction tuning. Otter's approach to in-context instruction tuning echoes and amplifies the image-text interleaved pretraining objectives common in multi-modality models, thereby enhancing the training convergence and augmenting model capacities.
    \item \textbf{MIMIC-IT Dataset:} To complement Otter's capabilities and advance multi-modality model research, we present the \textbf{MIMIC-IT} dataset. This comprehensive dataset encompasses over 3 million multimodal instruction-response pairs, including approximately 2.2 million unique instructions, covering a wide array of real-world images and videos.
    \item \textbf{Comprehensive Evaluations and Analysis:} Our analysis reveals significant benefits of multi-modal in-context instruction tuning using the MIMIC-IT dataset, enhancing Otter's training efficiency, generalization, and performance. MIMIC-IT's high-quality and diversity improves Otter's robustness in image and video evaluation benchmarks and its proficiency in diverse video and multi-image tasks, demonstrating its potential of serving as multi-mdoal general purpose assistant.
\end{itemize}

\section{Related Work}
\subsection{In-Context Learning}
In-context learning (ICL) in LLMs facilitates task adaptation through contextual input, leveraging inherent knowledge without the need for retraining~\cite{brown2020language,chowdhery2022palm}. The mechanics of ICL, however, remain underexplored. Studies suggest that LLMs implicitly undergo meta-learning during this process~\cite{akyurek2022learning,von2023transformers,garg2022can,dai2022can,min2022rethinking}, although evidence exists of performance resilience despite random label assignments in demonstrations. Instruction tuning improves the multi-tasking abilities of LLMs~\cite{sanh2021multitask,ouyang2022training,gpt4,flan,flant5,wang2022super}, yet it is unclear whether these capabilities are intrinsic to pretrained models or acquired during tuning. Research indicates that noisy instances generated by pretrained LLMs can be effectively utilized in instruction tuning, suggesting inherent connections between ICL and instruction tuning~\cite{wang2022super,honovich2022unnatural,ye2023context}. 

Recent studies have concentrated on instruction tuning in multimodal models, enabling them to process and respond to multi-modal instructions~\cite{llava,llama_adapater,instruct_blip,flamingo,awadalla2023openflamingo}. In this domain, the integration of ICL with instruction tuning remains unexplored except that Flamingo~\cite{flamingo} has exhibited robust few-shot ICL capabilities after pretraining on image-text interleaved webpage data. 

\subsection{Multi-modal Instruction Tuning Dataset}
The concept of instruction tuning in multi-modal models was first presented in Multi-Instruct~\cite{xu2022multiinstruct}, covering a broad spectrum of tasks involving visual comprehension and multi-modal reasoning, including Visual Question Answering~\cite{goyal2017making,zhu2016visual7w,vg}. Mini-GPT4~\cite{mini_gpt4} followed by creating an instruction-based dataset combining Conceptual Caption~\cite{cc3m,cc12m}, SBU~\cite{im2text}, and LAION~\cite{laion_400m} with custom instruction templates. More recently, LLaVA-Instruct-158K~\cite{llava} enhanced instruction tuning datasets by integrating self-instruct and ChatGPT~\cite{gpt4} methodologies with manually crafted seed instructions on COCO images~\cite{coco}. Follow up works extend in various aspects such as text richness~\cite{zhang2023llavar}, counter-examples~\cite{liu2023aligning}, multi-images~\cite{zhao2023mmicl}, larger-scale~\cite{zhao2023svit}, \textcolor{black}{3D modality~\cite{yin2023lamm}, fine-grained detail~\cite{chen2024sharegpt4v} and including videos~\cite{maaz2023videochatgpt}. }

\section{Otter Model}
\label{sec:otter_model}
The Otter model, inspired by the Flamingo framework~\cite{flamingo}, integrates pre-trained vision models (CLIP~\cite{clip}) and language models (\textit{e.g.} OPT~\cite{zhang2022opt}, LLaMA~\cite{llama}). Flamingo's training utilizes interleaved image-text data from web sources~\cite{zhu2023multimodal,common_crawl_nodate} and employs a perceiver resampler architecture for processing visual inputs. 
Despite Flamingo's unavailability, an opensourced replicate, the OpenFlamingo~\cite{awadalla2023openflamingo} project reproduced its architecture and training methods and provided pretrained model weights. 

\subsection{Architecture}
\vspace{-1mm}
We built Otter model based on a pretrained OpenFlamingo-9B. The crux of Otter's implementation\footnote{When unspecified, \textit{Otter} primarily refers to the Otter-9B model, with specific Image and Video versions indicated by additional suffixes.} is efficiently integration of two critical components: MPT-7B language decoder~\cite{MosaicML2023Introducing} and CLIP VIT-L/14 vision encoder~\cite{radford2021learning}. These two modules are connected via the perceiver resampler and the cross-gated attention modules as adopted in Flamingo-series models~\cite{flamingo}. Initially, the perceiver resampler module ingests a sequence of image or video features to produce a fixed set of visual tokens. Subsequently, these tokens condition the language layers through cross-gated attention modules, where these tokens act as keys and values while text from preceding layers serves as queries. Cross-gated attention enables that language layers is kept intact at initialization for improved stability and performance.

To optimize efficiency and accelerate experimental iterations, we only train the perceiver resampler and cross-gated attention modules within the pretrained OpenFlamingo-9B, collectively encompassing 1.4B parameters.

\subsection{Learning}
We train Otter on visual instruction tuning data, denoted as $\mathcal{D} = \{(\mathbf{X}_v, \mathbf{T}_i, \mathbf{T}_r)\}$ with token-level supervised fine-tuning~\cite{ouyang2022training,touvron2023llama}. During the training phase, the perceiver resampler processes images $\mathbf{X}_v$, converting them into visual tokens. These tokens are then mixed with the text modality within the language model's layers. These visual tokens condition subsequent layers via cross-gated attention modules. The training paradigm aligns with the next-token prediction objective, characteristic of the GPT series models~\cite{brown2020language,gpt4}, with the integration of visual and textual inputs. The likelihood of a targeted generated response $\mathbf{T}_{r}$ is formulated as:
\begin{equation}
p(\mathbf{T}_{r} \mid \mathbf{T}_{i}, \mathbf{X}_{v}) = \prod^{L}_{l=1} p(t_{l} \mid \mathbf{X}_{v}, \mathbf{T}_{i}, \mathbf{T}_{r, < l}).
\end{equation}
Here, $\mathbf{T}_{i}$ represents the instruction tokens, and $\mathbf{T}_{r, < l}$ signifies the sequence of response tokens preceding the current token $t_{l}$. In the inference stage, the language decoder's text tokenizer translates these tokens into coherent natural language.
Based on our preliminary experiments and the aim for training stability, Otter is primarily categorized into two variants based on the differing inputs of $\mathbf{X}_{v}$. The specifics of these two variants are detailed as follows:

\begin{figure}[htp]
    \centering
    \includegraphics[width=\columnwidth]{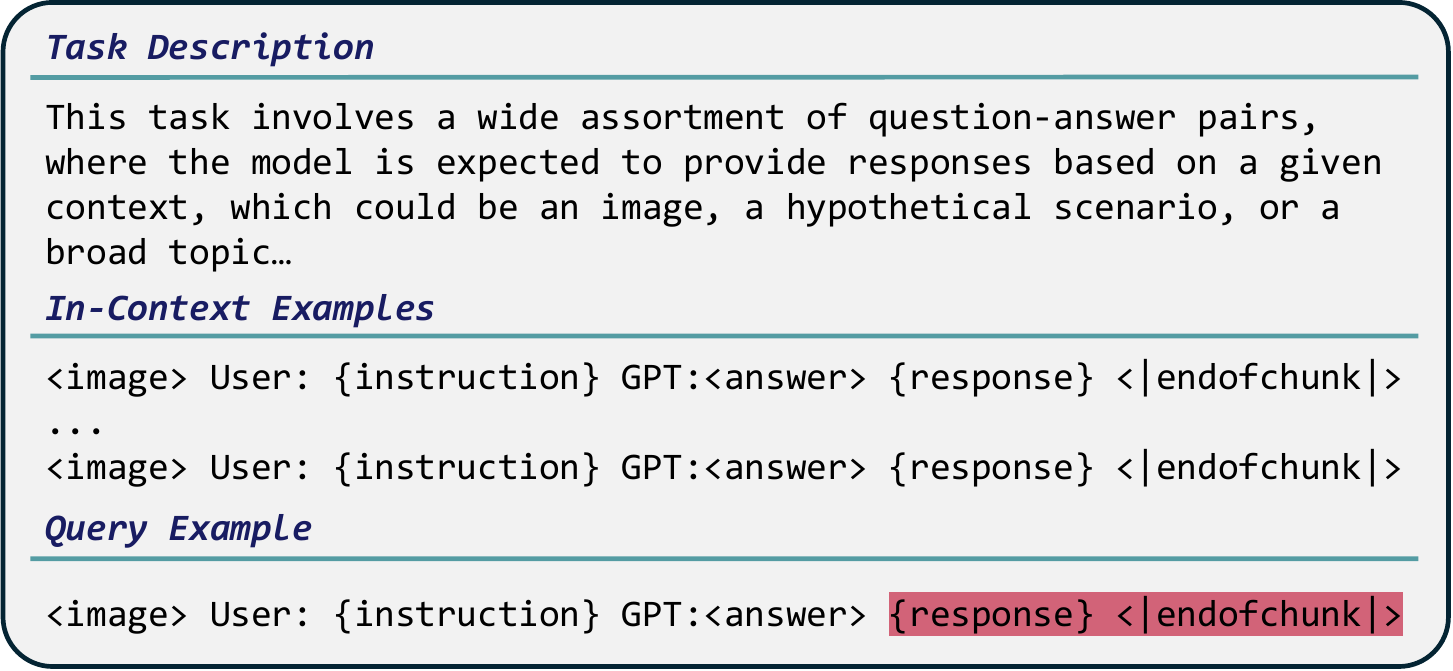}
    \caption{Instruction tuning template for the Otter model, where the targeted output in the decoding process is colored.}
    \label{fig:instrct_template}
\end{figure}

\noindent \textbf {Otter-Image} processes visual inputs $\mathbf{X}_{v}$ as tensors (\texttt{[N, T, C, H, W]}), where \texttt{N} is the number of images, \texttt{T} the frame count for videos, and \texttt{[C, H, W]} the typical image dimensions (\texttt{[3, 224, 224]}). It enriches in-context instruction tuning by using the \texttt{N} dimension for contextual images, aligning with instruction-response pairs (~\cref{fig:instrct_template}). The model utilizes \texttt{User} and \texttt{GPT} role labels for improved dialogue interaction.

\noindent \textbf  {Otter-Video} extends Otter-Image's functionality to dynamic content by leveraging the \texttt{T} dimension for sequential frame processing. This feature allows Otter-Video to interpret video inputs as temporally coherent frames with added time embeddings.

\section{MIMIC-IT Dataset}
\label{sec:data_format}
The MIMIC-IT dataset consists of visual instruction tuning data, represented as $\mathcal{D} = {(\mathbf{X}_v, \mathbf{T}_i, \mathbf{T}_r)}$. Each instance includes a set of $N$ images, forming a tuple: $(\mathbf{T}_i^q, \mathbf{T}_r^q, \mathbf{X}_v^q)$, where $\left\{\mathbf{x}_{j=1}^{N}\right\} \subseteq \mathbf{X}_v^q$. Here, $\mathbf{T}_i^q$ denotes the $q$-th instruction, $\mathbf{T}_r^q$ the response, and $\mathbf{X}_v^q$ the associated images or videos~\footnote{Videos are seen as sequential images.}. The primary objective within the Otter model framework (\cref{sec:otter_model}) is modeling the probability $p_{\theta}(\mathbf{T}_r^q \mid (\mathbf{T}_i^q, \mathbf{X}_v^q))$ with parameters $\theta$. The model is tasked with generating the response $\mathbf{T}_r^q$ for each query $(\mathbf{T}_i^q, \mathbf{X}_v^q)$, exemplifying the standard instruction tuning procedure in a visual language model.


A set of in-context examples is defined as $(\mathbf{T}_i^k, \mathbf{T}_r^k, \mathbf{X}_v^k)_{k=1}^{M}$, where $M$ is the count of such sets. The context function $C_{\psi}:(\mathbf{T}_i^q, \mathbf{X}_v^q) \mapsto {(\mathbf{T}_i^k, \mathbf{X}_v^k)}_{k=1}^{M}$ represents these in-context examples in relation to the current query. Consequently, the dataset format integrates query examples with corresponding in-context examples as follows:


\begin{equation}
d_q = (\mathbf{T}_i^q, \mathbf{T}_r^q, \mathbf{X}_v^q, C_{\psi}(\mathbf{T}_i^q, \mathbf{X}_v^q)), \quad d_q \sim \mathcal{D}_{\texttt{MIMIC-IT}}
\end{equation}

The visual language model, now incorporating in-context examples, is denoted as $p_{\theta}(\mathbf{T}_r^q | (\mathbf{T}_i^q, \mathbf{X}_v^q, C_{\psi}(\mathbf{T}_i^q, \mathbf{X}_v^q)))$. The context function $C_{\psi}$, being task-specific, necessitates different organizational strategies for in-context examples depending on the current query. 

The construction of the MIMIC-IT dataset entails extracting data from source images/videos and existing annotations, employing the self-instruct~\cite{self_instruct} approach. This process tasks ChatGPT with generating diverse questions based on the provided information and formulating its own responses. We guide ChatGPT to assume different roles in assorted scenarios using \texttt{System Messages} and \texttt{In-Context Examples}, creating data tailored to specific situations. 
We have an automatic pipeline, named \textbf{Syphus}, that assists us of this process. However, due to space constraints in the main paper, the intricacies of this method are detailed in the appendix. Additionally, the MIMIC-IT dataset utilizes various methods for constructing these examples or contextual information within its subsets, which will be further explicated in the subsequent section.

\begin{table*}[t]  
\tabstyle{4pt}
\caption{\textbf{Comparative Analysis of MIMIC-IT and Other Multi-Modal Instruction Datasets.} MIMIC-IT distinguishes itself with several notable characteristics: (1) Extensive size, encompassing a corpus of 3M instruction tuning examples specifically designed for LMMs; (2) Inclusion of video data; (3) Support for in-context instruction tuning; (4) Multilingual capabilities; (5) Data sourced from seven diverse contexts, including indoor and outdoor environments, TV dramas, and egocentric perspectives. In this table, \textit{lang.} refers to language, \textit{vis.} denotes vision, and \textit{uni. inst.} signifies unique instructions.}
\label{tab:comparison}
\renewcommand{\arraystretch}{1.1}
\begin{tabular}{lccccccc}
\arrayrulecolor{ntu_blue} \toprule
\textbf{Dataset} & \textbf{Source/Subset} & \textbf{In-Context Modality} & \textbf{Video} & \textbf{\#Clips/Images} & \textbf{\#Uni. Inst.} & \textbf{\#Instances} & \textbf{Lang.} \\ \arrayrulecolor{ntu_red} \shline
MiniGPT-4~\cite{mini_gpt4} & CC~\cite{changpinyo2021conceptual} & -/- & \textcolor{red}{\usym{2717}} & - / 134M & 4 & 5K  & English \\
LLaVA~\cite{llava} & COCO~\cite{coco}& lang./- & \textcolor{red}{\usym{2717}} & - / 81K & 156K & 156K & English \\ 
LLAVAR~\cite{zhang2023llavar} & LAION-5B~\cite{schuhmann2022laion} & lang./- & \textcolor{red}{\usym{2717}} & - / 16K & 16K & 16K & English \\ 
LRV~\cite{liu2023mitigating} & VisText~\cite{tang2023vistext} & lang./- & \textcolor{red}{\usym{2717}} & - / 35K & 400K & 400K & English \\ 
M3IT~\cite{li2023m} & Combined & lang./- & \textcolor{ForestGreen}{\usym{2713}} & 37.7K / 2.0M & 400 & 2.4M & Multi. \\ 
SVIT~\cite{zhao2023svit} & Visual Genome~\cite{krishna2017visual} & lang./- & \textcolor{red}{\usym{2717}} & - / 108.1K & 4.2M & 4.2M & English \\
Video-ChatGPT~\cite{maaz2023videochatgpt} & ActivityNet-200~\cite{heilbronactivitynet} & lang./- & \textcolor{ForestGreen}{\usym{2713}} & 13K / - & 45K & 100K & English \\ \shline
\multirow{9}{*}{\textbf{MIMIC-IT}} & LA-I2I/T2T & lang./vis. &  \textcolor{red}{\usym{2717}} & - / 81K & 261K & 156K & \multirow{8}{*}{\begin{tabular}[c]{@{}c@{}}\textbf{Multi.} \end{tabular}} \\
& GCD & lang./vis. &  \textcolor{red}{\usym{2717}} & - / 81K & 261K & 141K  & \\
& SD & lang./vis. &  \textcolor{red}{\usym{2717}} & - / 9K & 10K & 16K  & \\
& SN & lang./vis. &  \textcolor{red}{\usym{2717}} & - / 0.5K & 4.8K & 6.6K  &  \\
& DC & lang./vis. & \textcolor{ForestGreen}{\usym{2713}} & 16K / 1M & 40K & 63K  &  \\
& VST & lang./vis. & \textcolor{ForestGreen}{\usym{2713}} & - / 16K & 32K & 34K  & \\
& TVC & lang./vis. & \textcolor{ForestGreen}{\usym{2713}} & 86K / 577K & 86K & 89K  \\
& E4D & lang./vis. & \textcolor{ForestGreen}{\usym{2713}} & 400K / 6.4M & 1.9M & 2.5M  & \\
\cmidrule{2-7}
& Total & \textbf{lang./vis.} & \textcolor{ForestGreen}{\usym{2713}} & \textbf{502K / 8.1M} & \textbf{2.2M} & \textbf{3M} \\
\arrayrulecolor{ntu_blue} \bottomrule 
\end{tabular}  
\end{table*}

\subsection{Data Exploration}
\label{sec:data_source}
To enrich large multi-modal models, we developed eight diverse subsets within the MIMIC-IT dataset. These encompass scenarios like responding to instructions with examples (LA-T2T/I2I), analyzing image sequences (VST), and comparing images (GSD, SD). Additionally, they help in understanding TV show narratives (TVC) and enable a first-person view assistant (E4D) to interact using video inputs, improving user experience in daily tasks.

\noindent \textbf {LLaVA-I2I/T2T (LA-I2I/T2T)}
Since LLaVA-Instruct~\cite{llava} predominantly features single image-instruction-response pairs, for LA-I2I/T2T in MIMIC-IT, we aim to augment its utility for multi-modal models like Otter and Flamingo by generating in-context examples. We utilize CLIP similarity for instance matching with the following two approaches. For image-to-image matching within a MIMIC-IT dataset subset, we compute CLIP similarity scores and select the top-$K$ similar images. Similarly, for instruction-based matching, the top-$K$ examples are identified using text-to-text similarity. For a given query example $\mathbf{d}_q$, the in-context sets $\mathcal{D}_{\texttt{I2I}/\texttt{T2T}} = \{\mathbf{d}_1, \mathbf{d}_2, \ldots, \mathbf{d}_K\}$, crafted for LA-I2I and LA-T2T, are composed of individual examples $\mathbf{d}_i = (\mathbf{T}_i^c, \mathbf{T}_r^c, \mathbf{X}_v^c)$.

\noindent \textbf{General Scene Difference (GSD).}
Learning to discern differences between images is vital for understanding real-world changes. MIMIC-IT encompasses two kinds of discerning difference tasks. The first type, General Scene Difference, involves creating a pair of images by determining the most similar one to the current image, utilizing image-to-image CLIP similarity from the COCO~\cite{coco}. In GSD, we leverage the original image captions and object detection annotations to prompt ChatGPT with self-instruct strategy to generate instructions to ask about the difference between given two images.

\noindent \textbf{Subtle Difference (SD).}
Besides general difference, we also create Subtle Difference (SD), features pairs of similar images with subtle distinctions sourced from the Spot-the-Diff dataset\cite{spot_the_diff}, extracted from surveillance footage. We prompt ChatGPT to generate the instruction-response pairs according to original descriptions provided by dataset, focusing on identifying differences between the paired images.

\noindent \textbf{Visual Story Telling (VST).}
To enhance LMMs' context comprehension and narrative creation, we design the task Visual Story Telling, where source images and annotations are from~\cite{huang2016visual}, comprising sequential images of an event (\textit{e.g.} presidential selection) and corresponding inquiry questions. Annotations provide narratives and timelines not evident in the images alone. We direct ChatGPT to interpret these images and respond to related queries conversationally. The task also includes challenging questions to foster logical thinking and creativity. Notably, each instance comprises multiple image inputs and relevant instruction-response pairs, with prior conversations acting as contextual examples.

\noindent \textbf{Scene Navigation (SN).}
To demonstrate virtual assistants' planning skills, we use 2D photos of indoor scenes, derived from RGB-D images in ScanNetv2~\cite{scan_net}. We generate first-person view 2D representations of rooms and apply ScanRefer's visual annotations~\cite{chen2020scanrefer}. ChatGPT then formulates instructions for human interaction in these settings. The process involves ChatGPT creating a room owner persona and plans compatible with both the persona and room layout, ensuring context-sensitive user assistance. Like VST, SN instances comprise image sequences and corresponding instruction-response pairs.

\noindent \textbf{Dense Captions (DC).}
To enhance video comprehension, the DC subset integrates dense captions by Krishna et al.~\cite{krishna2017dense} with clips from longer videos, converted into 1FPS image frames. ChatGPT creates self-guided instructions based on questions covering video content aspects like visuals, human actions, event sequences, and causal links. Each DC instance includes sequential images with several instruction-response sets, serving as mutual in-context examples.

\noindent \textbf{TV Show Captions (TVC).}
To improve LMMs' social reasoning, TVC pairs TV show clips with captions, using resources from~\cite{lei2020tvr}. These clips facilitate the analysis of complex character interactions and social dynamics. LMMs are tasked with interpreting these narratives, investigating character motivations and relationships. This focus on social interactions and plot understanding prepares LMMs for real-world scenarios and varied user queries. Like VST, a TVC instance includes sequential images and associated instruction-response pairs, each serving as contextual examples for the others.

\noindent \textbf{Ego4D (E4D).}
E4D leverages egocentric video data to simulate the experience of Augmented Reality (AR) assistants interacting in real-life settings. Prompting ChatGPT for instructional generation based on visual cues, the aim is to mirror User/AR assistant interactions. The tasks and questions formulated are designed to elicit context-sensitive responses, enhancing the practicality of the LMMs in daily user-assistant interactions, underscoring the potential in providing actionable insights in day-to-day activities. The instance and in-context examples format is similar to DC and TVC.

\begin{table*}[tp]
\centering
\tabstyle{5pt}
\renewcommand{\arraystretch}{1.2}
\caption{Performance comparison of Otter-Image with recent open-sourced LMMs, detailing trainable parameters and instruction/response data pairs. The best is represented in \textbf{bold} while the second-best in \underline{underline}.}
\label{tab:image_benchmark_performance}
\resizebox{0.98\textwidth}{!}{%
\begin{tabular}{c|ccccccccccc}
\arrayrulecolor{ntu_blue}\toprule
\textbf{Model} & \textbf{Train Param.} & \textbf{POPE} & \textbf{MME$_{\texttt{Per.}}$} & \textbf{MME$_{\texttt{Cog.}}$} & \textbf{MMBench} & \textbf{MMMU$_{\texttt{val.}}$} & \textbf{SEEDBench} & \textbf{MM-Vet} & \textbf{MathVista} & \textbf{ScienceQA} \\ \arrayrulecolor{ntu_red} \shline
OpenFlamingo-9B~\cite{open_flamingo} & 1.3B & - & - & - & 5.7 & 28.7 & 24.8 & 24.8 & - & - \\
Idefics-9B$_{\texttt{Instruct}}$~\cite{laurenccon2023obelics} & 9.0B & 74.6 & 187.9 & 1351.8 & 45.5 & - & 44.5 & 23.7 & 19.8 & - \\
mPLUG-Owl$_{\texttt{V}}$~\cite{ye2023mplugowl} & 9.6B & - & - & - &  - & 32.7 & 34.0 & - & - & 37.9 \\
LLaMA-Adapter~\cite{zhang2023llama} & 1.8B & - & - & - & 39.5 & 29.8 & - & - & - & 85.2 \\
InstructBLIP-7B~\cite{dai2023instructblip} & 0.2B & - & - & - & 36.0 & - & 53.4 & 26.2 & - & 60.5 \\
InstructBLIP-13B~\cite{dai2023instructblip} & 0.2B & 78.9 & 291.8 & 1212.8 & 33.9 & - & - & 25.6 & 25.3 & 63.1 \\
Qwen-VL-7B$_{\texttt{Chat}}$~\cite{Qwen_VL} & 9.6B & - & \textbf{360.7} & 1487.5 & \underline{61.8} & 35.9 & 58.2 & - & - & \underline{68.2} \\
LLaVA-7B$_{\texttt{1.5}}$~\cite{liu2023improvedllava} & 7.0B & 85.9 & - & \underline{1510.7} & 59.5 & 29.2 & \textbf{58.6} & \underline{30.5} & - & 66.8 \\ \arrayrulecolor{ntu_red} \shline
\textbf{Otter-Image-9B} & 1.3B & \textbf{86.5} & \underline{332.5} & \textbf{1525.8} & \textbf{62.1} & 32.2 & \underline{57.5} & \textbf{30.6} & \textbf{27.5} & \textbf{69.3} \\ \arrayrulecolor{ntu_blue}\bottomrule
\end{tabular}
}
\end{table*}


\subsection{Dataset Statistics}
\begin{figure*}[t]
\centering
	\begin{tabular}{c c}
\includegraphics[height=5.0cm]{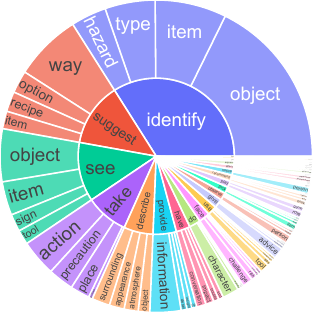} & 
\includegraphics[height=5.0cm]{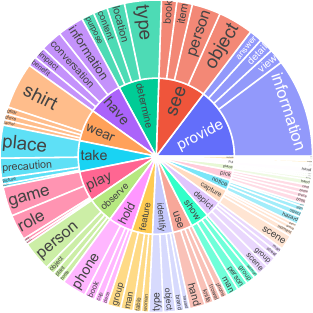}  \\
		(a) Instructions
		&
		(b) Responses \\

  \includegraphics[width=0.95\linewidth]{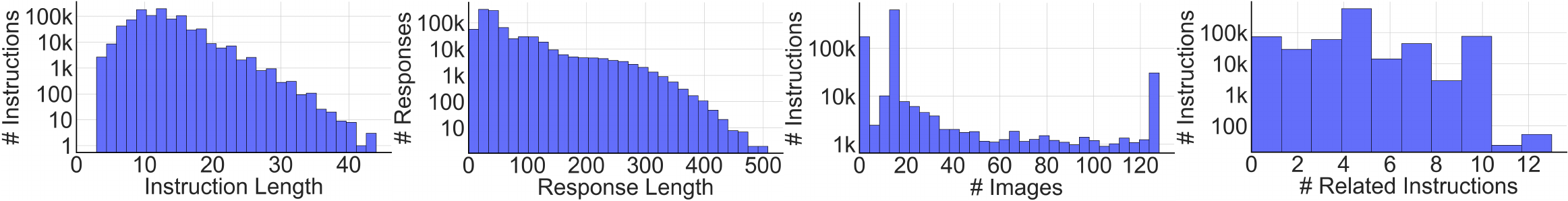}
  \hspace{-65mm}
  &  \\
  \hspace{-3mm}
  (c) Statistics of instructions and responses.
  \hspace{-65mm}
  & 
   \end{tabular}
    \caption{\textbf{The data statistics of multi-modal in-context instruction-response pairs.} (a) and (b), the root verb-noun pairs of instruction and responses, where the inner circle of the plot represents the root verb of the output response, and the outer circle represents the direct nouns.
    (c) Statistics of instructions and responses, retaining 25\% of Ego4D instructions for a more balanced distribution. \# Related instructions denotes the number of related instructions in an instance, given the same set of visual input data.}
    \label{fig:data_stats}
\end{figure*}

To examine the characteristics and diversity of the instructions (refer to ~\cref{fig:data_stats} (a)) and responses (refer to ~\cref{fig:data_stats} (b)), we analyze the verb-noun structure present in them, refering to~\cite{self_instruct}. Specifically, we employ spaCy for parsing the instructions, extracting the verb closest to the root, and retrieving its first direct noun object\footnote{\url{https://github.com/explosion/spacy-models/releases/tag/en_core_web_md-3.5.0}}. We plot the top 20 most frequently occurring root verbs alongside their top 4 direct noun objects. Our findings reveal that the sentence structure of responses exhibits greater diversity compared to that of instructions. Moreover, we demonstrate diversity in terms of the length of instructions/responses, the number of images per instruction, and the number of in-context examples per instruction, as depicted in ~\cref{fig:data_stats} (c).

\subsection{Safety and Ethical Considerations}
In the process of curating the MIMIC-IT dataset, several ethical, privacy, and potential bias concerns arise.

\noindent \textbf{Personally Identifiable Information} 
The MIMIC-IT dataset, primarily sourced from public datasets, has the potential to contain personally identifiable information (PII). The image and video sources from COCO, DC, and VIST are derived from online web searches. Consequently, the foundational dataset creators do not bear responsibility for personal information protection. In contrast, the TVC dataset comprises scenes from TV series. The E4D videos are either captured with informed consent in controlled environments or in public spaces, with faces and other PII obscured.

In essence, while MIMIC-IT may contain PII, it is primarily sourced from the public domain or obtained with consent, addressing concerns related to personal data protection.

\noindent \textbf{Gender and Race Ratio Distribution} 
In addressing potential ethical biases in image and video sources, we conducted a race and gender classification on our source images to assess their distribution. These findings, detailed in our paper, serve as guidelines for researchers utilizing our dataset and are presented in Table~\ref{tab:gender_race_distribution}.

Specifically, we sampled a uniform 5\% from each subset, proportional to the entire dataset's volume. For each image and video frame, we employed the model from~\cite{hyeon_face_classification} for facial recognition and race/gender classification. The classification labels covered combinations of {Asian, White, Black} $\times$ {Man, Woman}. Images with multiple faces were counted accordingly; those without faces were excluded. Our race/gender classification results for the sampled data are outlined in Table~\ref{tab:gender_race_distribution}.

\begin{table}[t]
\caption{\textbf{Race and Gender} Distribution in MIMIC-IT.}
\label{tab:gender_race_distribution}
\centering
\tabstyle{6pt}
\renewcommand{\arraystretch}{1.4}
\centering
\resizebox{0.8\columnwidth}{!}{%
\begin{tabular}{r|cccc}
\shline
\diagbox{\textbf{Gender}}{\textbf{Race}} & \textbf{Asia} & \textbf{White} & \textbf{Black} & \textbf{Total} \\ \midrule
\textbf{Woman}             & 17.3\%       & 19.5\%        & 9.7\%         & 46.5\%        \\ \hline
\textbf{Man}               & 18.4\%       & 22.5\%        & 12.6\%        & 53.5\%        \\ \hline
\textbf{Total}             & 35.7\%       & 42.0\%        & 22.3\%        & 100.0\%       \\ \bottomrule
\end{tabular}
}
\end{table}

To provide context, we also present reference distributions for race and gender, selecting the United States as a comparative framework due to its inclusive racial diversity, offering more balanced proportions than some Asian or European countries. The racial and gender compositions in the United States are presented in Tables~\ref{tab:us_race_distribution} and~\ref{tab:us_gender_distribution}, respectively.

\begin{table}[t]
\caption{\textbf{Racial Composition} in the United States.}
\label{tab:us_race_distribution}
\centering
\tabstyle{4pt}
\renewcommand{\arraystretch}{1.4}
\centering
\resizebox{0.9\columnwidth}{!}{%
\begin{tabular}{cccccccc|c}
\toprule
\textbf{Location} & \textbf{White} & \textbf{Black} & \textbf{Hispanic} & \textbf{Asian} & \textbf{Others} & \textbf{Total} \\ \hline
United States     & 58.2\%         & 11.6\%         & 19.0\%            & 5.7\% & 5.6\% & 100.0\%        \\ \bottomrule
\end{tabular}
}
\end{table}

The gender composition for reference in the United States is detailed in Table~\ref{tab:us_gender_distribution}.

\begin{table}[t]
\caption{\textbf{Gender Composition} in the United States}
\label{tab:us_gender_distribution}
\centering
\tabstyle{6pt}
\renewcommand{\arraystretch}{1.4}
\resizebox{0.8\columnwidth}{!}{%
\begin{tabular}{c|cc}
\toprule
\textbf{Location}      & \textbf{Biological Woman} & \textbf{Biological Man} \\ \midrule
United States & 49.28\%          & 50.72\%        \\ \bottomrule
\end{tabular}
}
\end{table}

\noindent \textbf{Satety Alignment in Instruction-Response Pairs} 
In our utilization of ChatGPT for generating instruction-response pairs, we stringently adhere to Microsoft Azure's ChatGPT\footnote{\url{https://azure.microsoft.com/en-in/products/ai-services/openai-service/}} content policy. This policy is meticulously crafted to ensure the generation of content that is both safe and ethical. It serves as a pivotal measure in mitigating potential biases, stereotypes, explicit content, or misinformation, thus aligning with broader initiatives in AI ethics and safety. By implementing ChatGPT's rigorous safety and alignment strategies, we not only uphold high standards of content safety but also actively contribute to the responsible development and deployment of AI technologies. This dedication is consistently demonstrated in the quality and reliability of the content produced by ChatGPT for our datasets.

\section{ChatGPT-Assisted Dataset Generation}
\label{app:syphus}

\begin{figure*}[htp]
    \centering
    \includegraphics[width=0.9\textwidth]{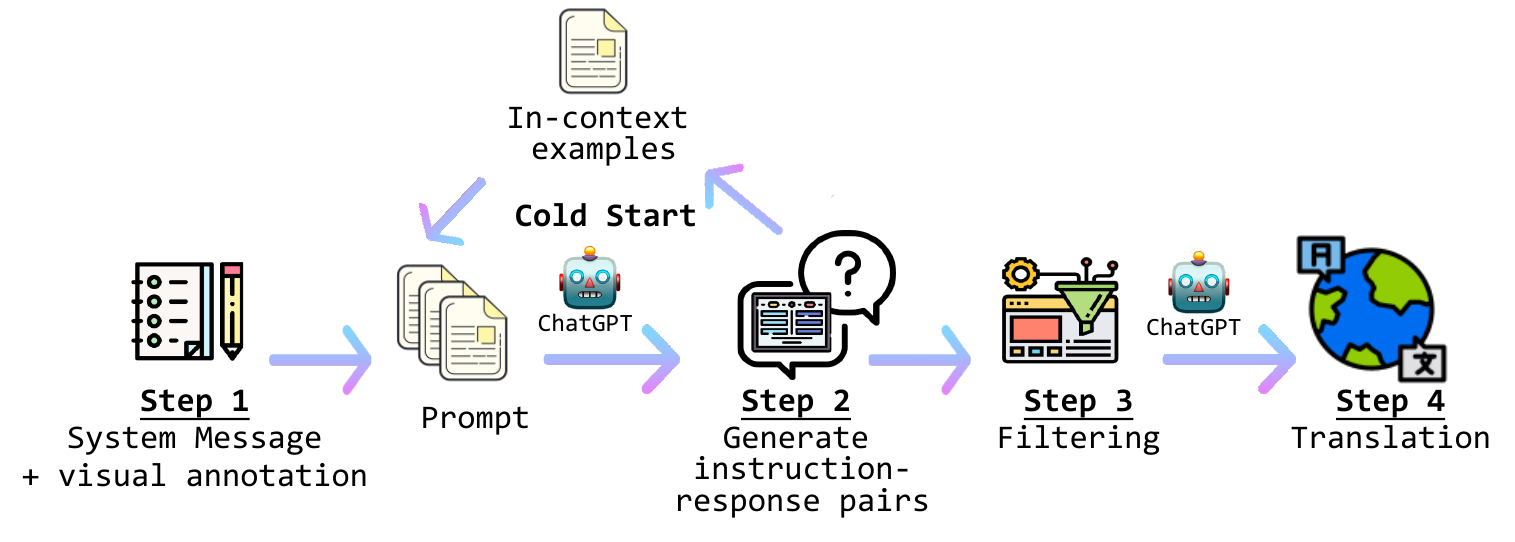}
    \caption{\textbf{Syphus overview.} We employ a cold-start stage to identify the optimal system message and in-context example for querying instruction-response pairs in a given dataset. Subsequently, Syphus, spanning steps 1 to 4, generates high-quality instruction-response pairs in eight languages.}
    \label{fig:Syphus}
\end{figure*}

\subsection{Generation Pipeline}
\textcolor{black}{We present \textbf{Syphus} (see~\Cref{fig:Syphus}), an automated pipeline for generating high-quality instruction-response pairs in multiple languages. Building upon the framework proposed by LLaVA~\cite{llava}, we utilize ChatGPT to generate instruction-response pairs based on visual content and original annotations.}

\textcolor{black}{To ensure the quality of the generated instruction-response pairs, our pipeline incorporates system messages, visual annotations, and in-context examples as prompts for ChatGPT. System messages define the desired tone and style of the generated instruction-response pairs, while visual annotations provide essential image information such as bounding boxes and image descriptions. In-context examples assist ChatGPT in learning within the context. Since the quality of coreset impacts subsequent data collection process~\cite{chen2022making}, we employ a cold-start strategy to enhance in-context examples before the large-scale query.  }

\textcolor{black}{During the cold-start stage, in-context examples are collected by prompting ChatGPT solely through system messages and visual annotations, employing a heuristic approach. This stage concludes only when satisfactory in-context examples are identified. In step 4, once the instruction-response pairs are obtained, the pipeline expands them into Chinese (zh), Japanese (ja), Spanish (es), German (de), French (fr), Korean (ko), and Arabic (ar). }


\subsection{Application Scenarios Demonstration}
\begin{figure*}[h]
    \centering
    \includegraphics[width=0.8\textwidth]{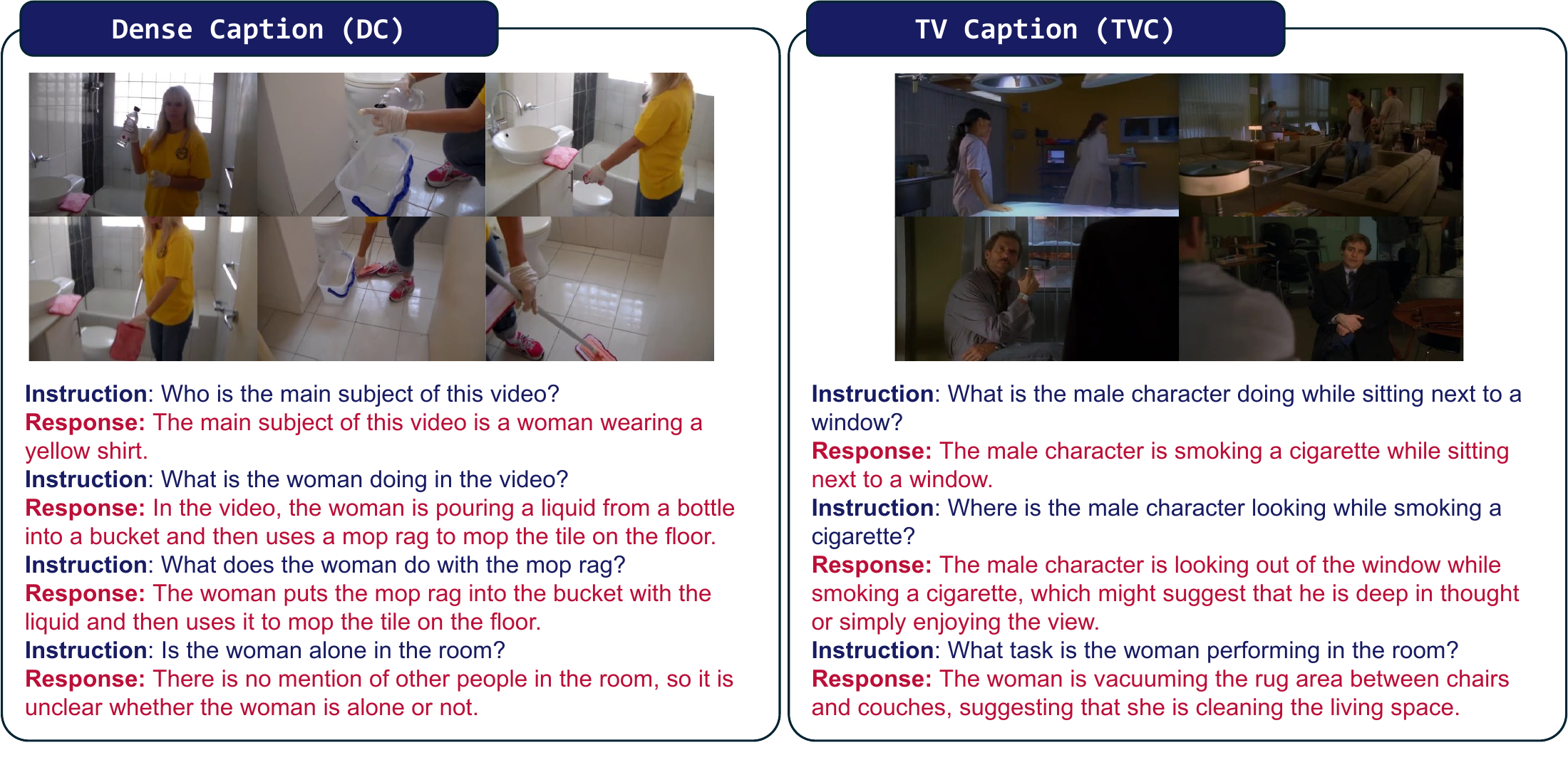}
    \caption{Demonstration examples in DC and TVC.}
    \label{fig:app_demo1}
\end{figure*}

\begin{figure*}[h]
    \centering
    \includegraphics[width=0.8\textwidth]{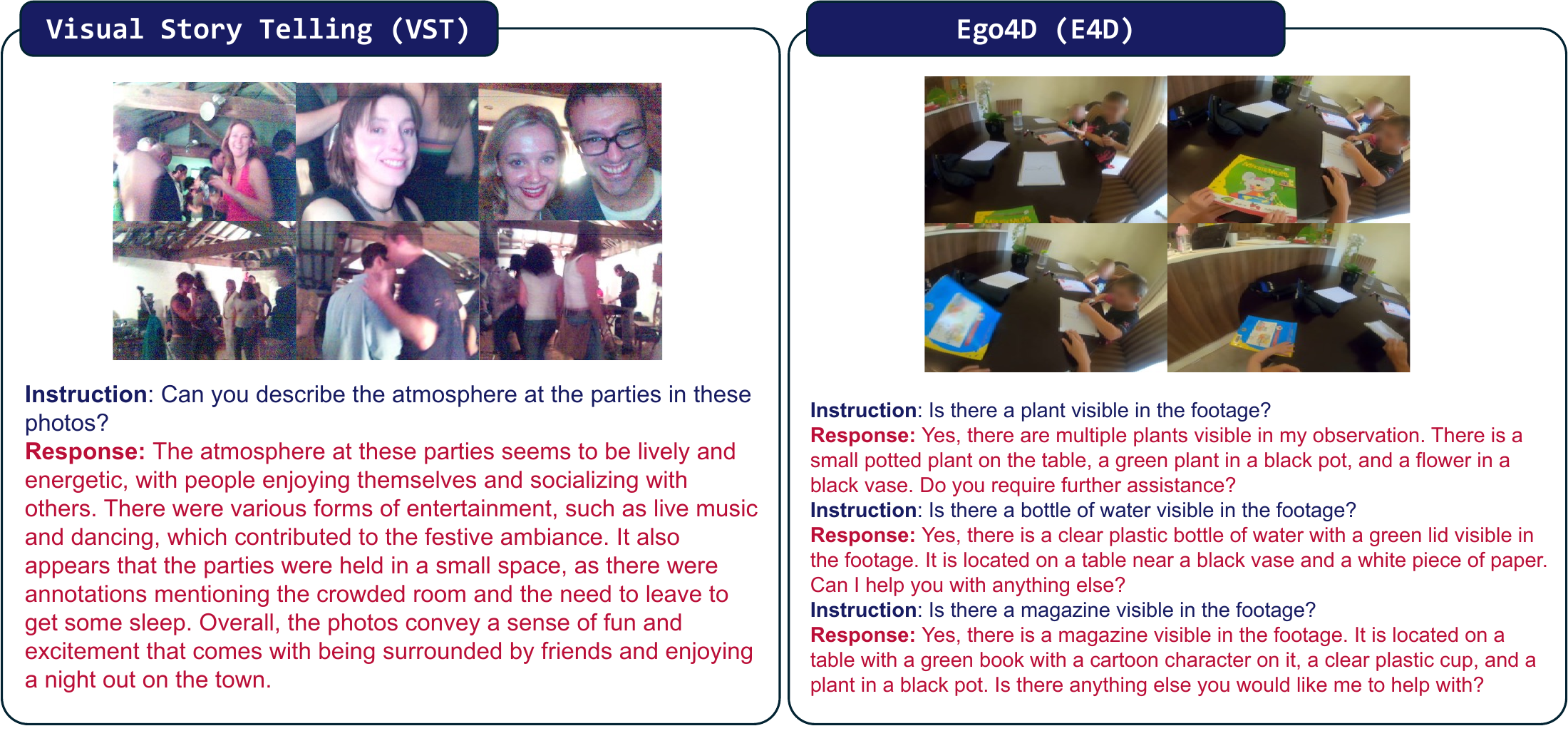}
    \caption{Demonstration examples in VST and E4D.}
    \label{fig:app_demo2}
\end{figure*}

\begin{figure*}[h]
    \centering
    \includegraphics[width=0.8\textwidth]{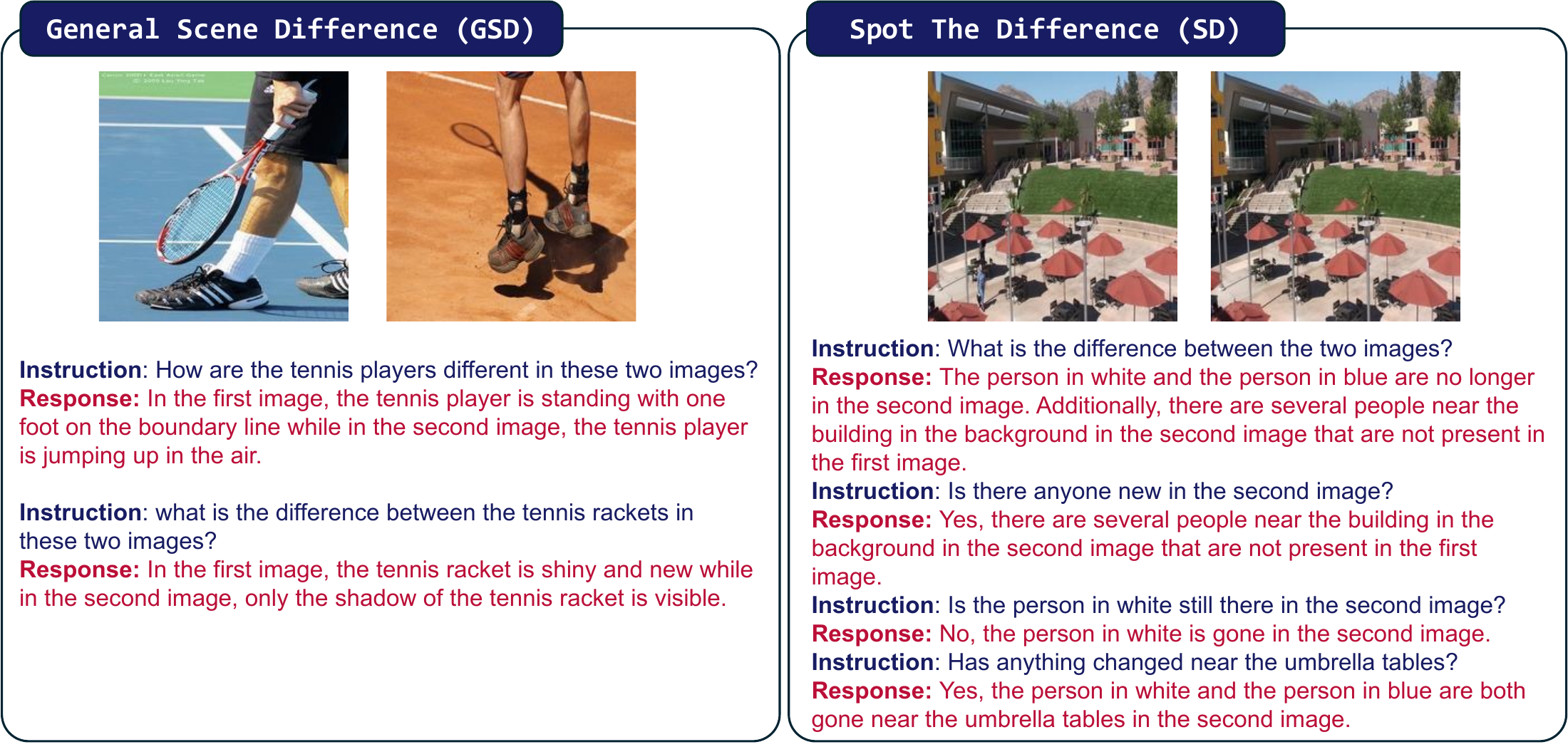}
    \caption{Demonstration examples in GSD and SD.}
    \label{fig:app_demo3}
\end{figure*}

\begin{figure*}[h]
    \centering
    \includegraphics[width=0.8\textwidth]{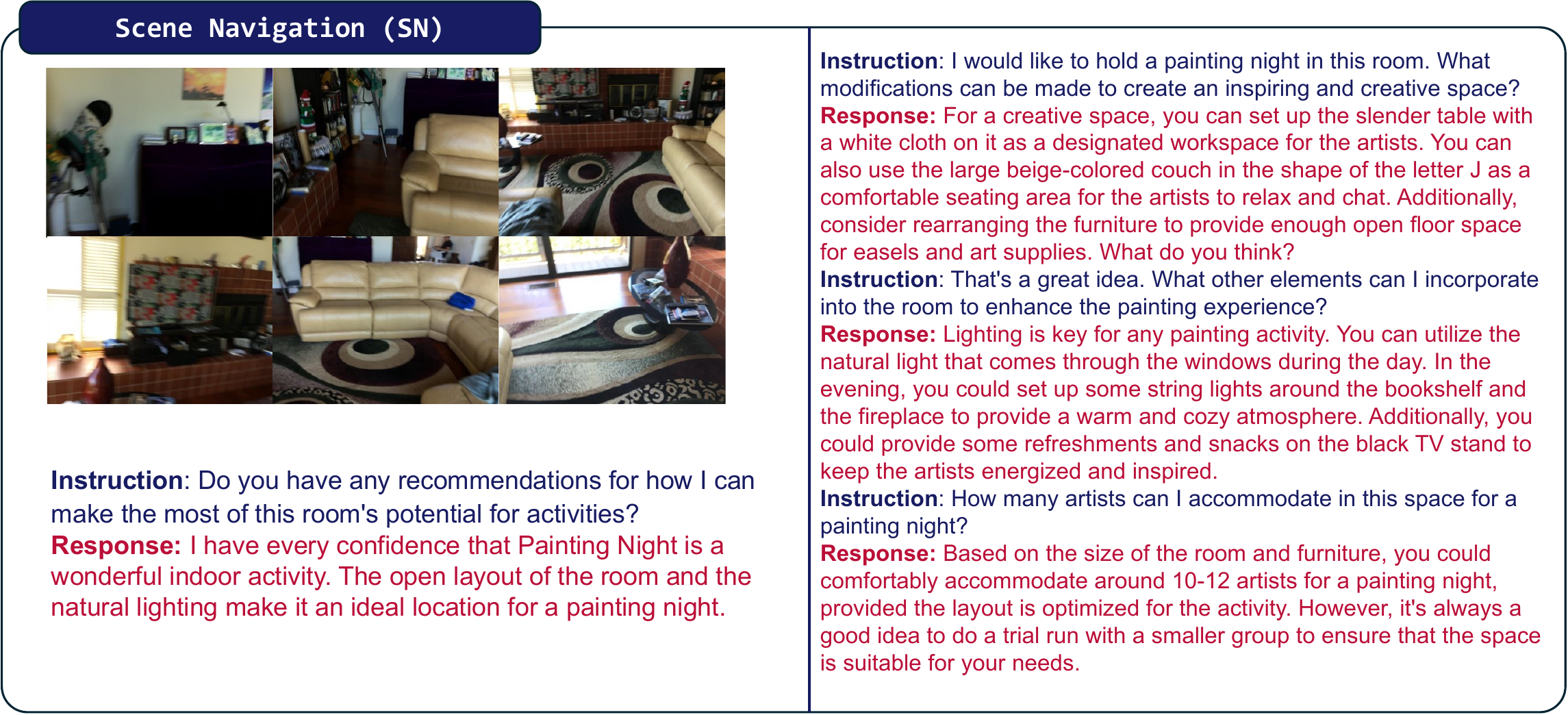}
    \caption{Demonstration examples in SN.}
    \label{fig:app_demo4}
\end{figure*}

\textcolor{black}{In~\cref{fig:app_demo1,fig:app_demo2,fig:app_demo3,fig:app_demo4}, we present a diverse array of sample scenarios selected from various subsets of our dataset. These samples vividly illustrate the versatility and adaptability of our model across a range of contexts and dialogues pertinent to specific scenarios.}

\textcolor{black}{Each scenario is carefully chosen to represent the distinct characteristics of its respective subset, demonstrating not only the breadth of our dataset but also the depth of the model's contextual understanding and response accuracy. These examples collectively showcase the model's robustness in interpreting and responding to a wide spectrum of dialogues and scenarios, an essential attribute for practical applications in diverse real-world settings.}

\subsection{Dataset Cleaning}
\textcolor{black}{After obtaining the instruction-response pairs generated by ChatGPT, the majority of the data satisfies our requirements, adhering to the format, content, style, and pronoun usage provided in the in-context examples. However, some data deviates from our specified pattern due to unforeseen circumstances. For instance, in the E4D dataset, we expect the ChatGPT-generated response to adopt a style in which the AI, acting as a first-person AR headset assistant, helps users solve problems. Consequently, the questions should be phrased in the second person, simulating a user's inquiry, while the responses should be in the first person, simulating the model's reply to the user's question. ~\cref{fig:lst_example} demonstrates an example of a correct dialogue.}

\begin{table}[t]
\begin{lstlisting}[style=instructformat]
User: Can the black and white snowboard be used for beginners?
GPT: Based on my observation, there is a black and white snowboard visible, but I cannot determine its suitability for beginners. Snowboarding can be a challenging sport, and it's essential to choose equipment that is appropriate for your skill level. Would you like me to find more information on snowboarding equipment?
\end{lstlisting}
\caption{Example of a correct dialogue.}
\label{fig:lst_example}
\end{table}

   
   





\textcolor{black}{
Nevertheless, the content generated by ChatGPT may occasionally contain errors, such as including \texttt{the user (you)} as shown in~\cref{fig:wrong_dialogue}, which might result from the information provided in our query.}

\begin{table}[!ht]
\begin{lstlisting}[style=instructformat]
User: Is there any equipment in the garden that I could use for digging?
GPT: According to what I saw, the user (you) picked up a garden pick mattock and a sledgehammer. These could be used for digging, but I cannot guarantee their availability or suitability for your needs. Is there anything else you need help with?
\end{lstlisting}
\caption{Example of a wrong dialogue.}
\label{fig:wrong_dialogue}
\end{table}


\textcolor{black}{We cross-checked each dataset with two human individuals, each person spent 15 hours to manually check the error patterns in different datasets, and, for fixable errors (such as the one above, where \texttt{user (you)} needs to be replaced with \texttt{you}), employed regular expressions to match and replace these errors with the correct expressions. For other errors that cannot be rectified en masse, we documented their locations and stored them in a dedicated \texttt{invalid.json} file. During the dataset iteration process, these samples were fed into ChatGPT to repeat the query and obtain accurate and usable data.}

\section{Experiments}
\label{sec:experiments}

\subsection{Data Mixture}
\label{subsec:data_mixture}
Consistent with the observations made in LLaVA-1.5~\cite{liu2023improvedllava}, we found that precisely controlling the response format in instruction tuning data significantly enhances benchmark performance. This improvement is particularly notable in tasks~\cite{fu2023mme}, which require binary \textit{yes}/\textit{no} answers. In addition to MIMIC-IT, our training dataset for Otter also includes selections from academic sources to further refine its capabilities. Data from these academic sources were predominantly sourced from the M3IT Collection\cite{li2023m}. For all datasets, Otter was exclusively trained on the \texttt{train} sets, enabling a focused evaluation on the \texttt{validation} and \texttt{test} sets.

For Otter-Image model, we compiled a data mixture from the following public datasets: VG~\cite{krishna2016visual}, VQAv2~\cite{antol2015vqa}, GQA~\cite{hudson2019gqa}, OKVQA~\cite{marino2019ok}, OCRVQA~\cite{mishraICDAR19}, A-OKVQA~\cite{schwenk2022okvqa}, TextQA~\cite{singh2019towards}, RefCOCO~\cite{yu2016modeling}, COCO-ITM~\cite{li2023blip}, ImageNet~\cite{deng2009imagenet}, ImageNet-Paragraphs~\cite{krause2016paragraphs}. For these datasets, we only construct the \textit{Answer Format} part in task description as the prefix for each instruction. 

For Otter-Video model, we trained it on the DC/TVC subsets from MIMIC-IT, with each video was initially sampled and stored into the dataset at 1 FPS. During batch-wise training, we uniformly extracted 64 frames for training (if a video had fewer frames, adjacent frames were repetitively sampled). We also used instruction tuning data from MSVD~\cite{chen2015microsoft} and MSRVTT~\cite{xu2017video}, where training data from these sources were limited to only 8 frames per clip.

\subsection{Benchmark Performance}
\label{subsec:benchmark}

\noindent \textbf{Image Benchmarks} In~\cref{tab:image_benchmark_performance}, we present a comprehensive comparison between Otter-Image and other state-of-the-art LMMs across a variety of benchmarks. We present performance in accuracy on benchmarks including POPE~\cite{li2023evaluating}, MM-Vet~\cite{yu2023mm}, MMBench~\cite{liu2023mmbench}, MathVista~\cite{lu2023mathvista}. On MMBench, we report results on test set. For MME~\cite{fu2023mme}, we report the aggregated scores in cognitive and perception to follow its evaluation convention. 

\noindent \textbf{Video Benchmarks} In \cref{tab:video_benchmark_performance}, we present a detailed quantitative assessment across several leading open-ended question-answer video datasets: MSRVTT-QA~\cite{xu2017video}, MSVD-QA~\cite{chen2015microsoft}, and ActivityNet-QA~\cite{yu2019activitynetqa}. This evaluation employs a zero-shot approach using ChatGPT-assisted scores to evaluate model proficiency. The focus is on measuring the precision of the model’s predictive responses. Additionally, we include a few-shot evaluation results of MSRVTT-Caption, as well as the evaluation details in the appendix for comprehensive coverage.

\begin{table}[tp]
\centering
\tabstyle{2pt}
\renewcommand{\arraystretch}{1.4}
\caption{Performance comparison of Otter-Video with recent open-sourced LMMs on video question-answer tasks. The best is represented in \textbf{bold} while the second-best in \underline{underline}.}
\label{tab:video_benchmark_performance}
\resizebox{\columnwidth}{!}{%
\begin{tabular}{r|ccc}
\textbf{Model} & \textbf{MSVD-QA}~$\uparrow$ & \textbf{MSRVTT-QA}~$\uparrow$ & \textbf{ActivityNet-QA}~$\uparrow$ \\ \arrayrulecolor{ntu_red} \shline
LLaMA-Adapter$_{\texttt{V}}$~\cite{zhang2023llamaadapter} & 32.6 & 35.7 & 30.5 \\
Video-ChatGPT~\cite{maaz2023videochatgpt} & 49.1  & 48.7 & 47.2 \\
VideoChat~\cite{li2023videochat} & 47.6 & 48.0 & 48.4 \\
mPLUG-Owl$_{\texttt{V}}$~\cite{ye2023mplugowl} & \underline{64.1} & \underline{66.2} & \underline{47.4} \\ \arrayrulecolor{ntu_red} \shline
\textbf{Otter-Video} & \textbf{77.9} & \textbf{68.4} & \textbf{68.5} 
\end{tabular}
}
\end{table}

\subsection{Benefiting from Contextual Information}
\label{subsec:analysis}

From the above-described instruction template, it is evident that the multi-modal in-context instruction tuning (ICIT) process involves incorporating additional contextual information, namely \textit{task description} and \textit{in-context examples}, before each query example. This strategy, when compared to other instruction tuning data approaches, potentially enriches the context for the current query example and aids in steering the model towards generating the targeted response. To verify this intuition, we conducted a series of ablation studies on the effect of adding contextual information in following sections.

\noindent \textbf{Task Description}
For efficiency, our ablation study only uses the LLaVA-Instruct-158K dataset~\cite{llava}, categorizing it into three subsets: Detailed Description, Conversation, and Complex Reasoning. The task descriptions on each subset are constructed from three dimensions: (1) \textit{Question Coverage}. (2) \textit{Model Ability}. (3) \textit{Answer Format}. The three parts would prepend to each instruction as prefix to aid model better comprehend the instructions.

\begin{figure}[tp]
    \centering
    \includegraphics[width=0.80\columnwidth]{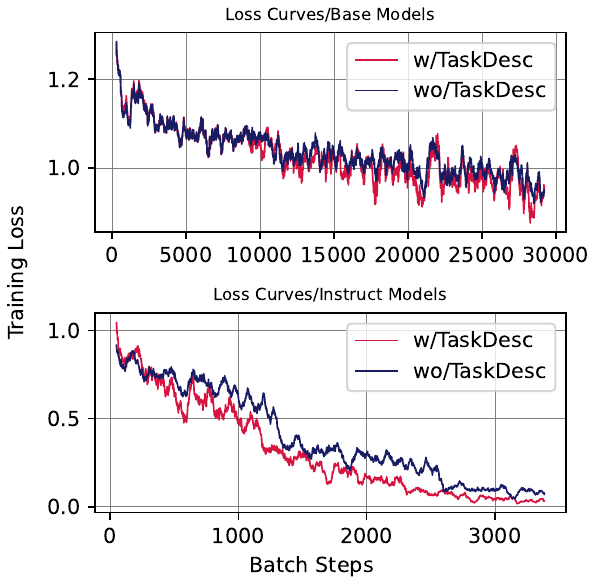}
    \caption{Comparative analysis of training convergence with and without task descriptions.}
    \label{fig:loss_task_desc}
\end{figure}

\cref{fig:loss_task_desc} compares model convergence with and without task descriptions during training. The figure is split into upper and lower sections: the upper shows the loss curve during finetuning from an OpenFlamingo base model, with a focus on the impact of task descriptions. Initially, both methods show similar loss, but after 20K steps, they begin to diverge. Our initial hypothesis was that training loss differences would be minimal. However, an experimental pivot led us to further explore this using the better-performing 'instruct models' from the first training phase. This revealed a growing loss divergence over time, with task description-inclusive models showing lower loss and improved convergence.

We also present ablation results across various benchmarks, focusing on response format optimization for benchmark performance. To aid the model learn to respond in certain answer format, two subsets were constructed: one (830 examples) from COCO~\cite{coco} for object presence queries (\textit{Yes}/\textit{No} responses), and another (2000 examples) from RefCOCO~\cite{kazemzadeh2014referitgame} for identifying objects from options. Both subsets were created through rule-based construction from annotations or direct sampling from data samples.

\begin{table}[tp]
\tabstyle{6pt}
\renewcommand{\arraystretch}{1.2}
\caption{Performance gains through task description enhancements.}
\label{task_description_ablation}
\resizebox{0.35\textwidth}{!}{%
\begin{tabular}{r|cc}
\textbf{Method} & \textbf{MME-Cog}~$\uparrow$ & \textbf{MM-Vet}~$\uparrow$ \\ \arrayrulecolor{ntu_red} \shline
\color{gray}{Base Model} & \color{gray}{1212.9} & \color{gray}{23.7} \\ \hline
+ Question Coverage & 1315.2 & 24.5 \\
+ Model Ability & 1329.3 & 25.8 \\
+ Answer Format & 1414.5 & 22.3 \\ \hline
\textit{Eval} w/Full Desc. & \textbf{1415.7} & \textbf{27.0}
\end{tabular}}
\end{table}

We selected two benchmarks with different objectives for evaluation. MME-Cog is the Cognition subset of MME, featuring questions about art, celebrities, locations, \textit{etc.}, with answers limited to \textit{yes} or \textit{no}. MM-Vet focuses more on logical reasoning and basic mathematics, with a \textit{freeform} answer format. The correctness of responses is assessed using GPT-4. The model's ablation results on these benchmarks are presented in~\cref{task_description_ablation}. It is evident that controlling the \textit{Answer Format} through task description during training particularly enhances performance on MME~{\footnotesize(1212.9 $\rightarrow$ 1414.5)}. Such effect is not observed in MM-Vet. However, the most significant improvement in MM-Vet is derived from specifying the capabilities on which the model relies for answering questions, with a notable increase~{\footnotesize(23.7 $\rightarrow$ 25.8)}.

\begin{figure}[tp]
    \centering
    \includegraphics[width=0.80\columnwidth]{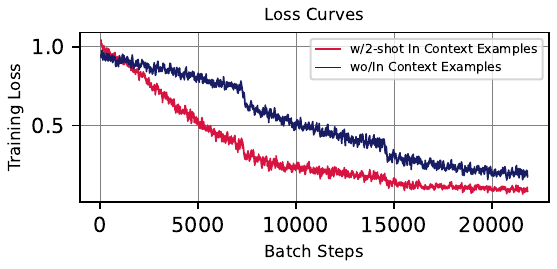}
    \caption{Comparative analysis of training convergence with and without 2-shot in-context examples, illustrating the impact on the model's learning trajectory.}
    \label{fig:loss_in_context}
\end{figure}

\noindent \textbf{In-Context Examples} As detailed in~\cref{sec:data_format}, apart from task descriptions, we incorporate contextual information through retrieval-based in-context examples. In our ablation study, we employed LA-Interleaved subsets from the MIMIC-IT dataset, with in-context examples retrieved using both text-to-text (T2T) and image-to-image (I2I) similarity matching. Our findings, presented in~\cref{fig:loss_in_context}, highlight the variations in convergence speed when employing 2-shot in-context examples.

\begin{table}[tp]
\tabstyle{7pt}
\renewcommand{\arraystretch}{1.3}
\caption{Ablations of in-context examples on MME-Cog.}
\label{tab:task_perf_ablations}
\resizebox{0.35\textwidth}{!}{%
\begin{tabular}{r|ccc}
\multirow{2}{*}{\textbf{Method}} & \multicolumn{3}{c}{\textbf{MME-Cog}~$\uparrow$} \\ 
 & 0 & 2 & 4 \\ \arrayrulecolor{ntu_red} \shline
\color{gray}{\textit{Train} wo/ICIT} & \color{gray}{1212.9} & \color{gray}{1310.7} & \color{gray}{1173.5} \\ \hline
\textit{Train} w/2-shot & 1227.7 & \textbf{1430.3} & 1397.8 \\
\textit{Train} w/4-shot & \textbf{1277.8} & 1387.9 & \textbf{1440.8}
\end{tabular}
}
\end{table}

We further demonstrate the performance improvements on benchmarks achieved through \textit{in-context instruction tuning}, as shown in Table~\ref{tab:task_perf_ablations}. The columns labeled $0$, $2$, and $4$ shots in the table correspond to the number of in-context examples included during the evaluation phase. To circumvent information leakage in the MME test set, we crafted in-context examples from web images, aligning the instructions and answers with the dataset's required format. The rows in the table represent the varying quantities of in-context examples used during the instruction tuning process.

\begin{figure}[htp]
\centering
\includegraphics[width=0.9\columnwidth]{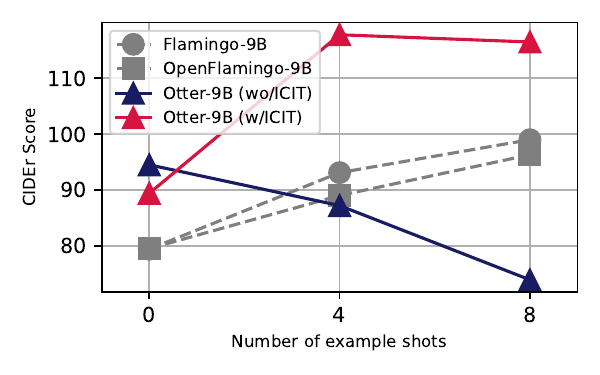}
\caption{Few-shot performance on COCO Caption.}
\label{fig:fewsot_coco}
\end{figure}


In few-shot evaluations, base models like Flamingo and OpenFlamingo, without instruction tuning, often show improved performance with more in-context examples due to ICL capabilities. Our ablation study, illustrated in~\cref{fig:fewsot_coco}, explores this phenomenon in multimodal instruct models. Notably, Otter (wo/ICIT) excel in 0-shot scenarios but falter as examples increase, implying a diminished ICL capacity when fine-tuned on single image-instruction-response pairs. Conversely, Otter (w/ICIT) markedly improve with more shots, underscoring ICIT's importance in preserving ICL effectiveness in multimodal settings. More few-shot evaluation results are provided in appendix.

To clarify how ICIT impacts model behavior,~\cref{fig:ict_demo} demonstrates how referencing in-context examples influences the trained model's text formatting control.

\begin{figure}[tp]
\centering
\includegraphics[width=\columnwidth]{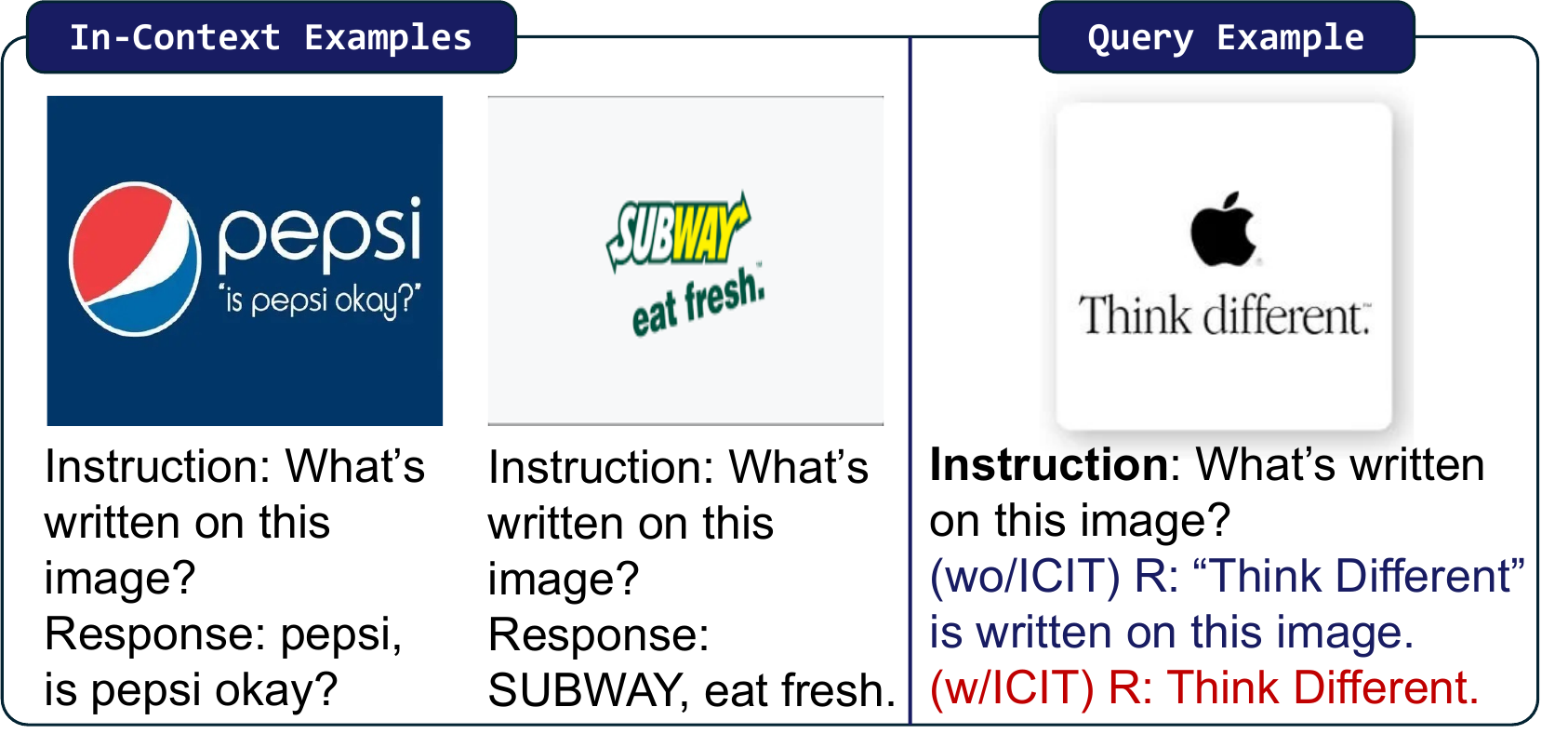}
\caption{Comparative demonstrations of Otter with and without in-context instruction tuning (ICIT).}
\label{fig:ict_demo}
\end{figure}

\subsection{Diverse Applications}
\label{subsec:application_scenarios}
MIMIC-IT extends Otter's capabilities beyond image understanding to diverse applications.

\noindent\textbf{Video Captioner and Assistant.} MIMIC-IT's extensive video collection, featuring dialogues on content and character relationships, enriches Otter's functionality. This includes YouTube content and series like \textit{Friends} in the DC and TVC subsets. Trained models act as video assistants, enhancing user interaction and enabling detailed video content annotation for research.

\noindent\textbf{Egocentric Visual Assistant.} Additionally, MIMIC-IT's egocentric videos and images, particularly from SN and E4D scenarios, focus on indoor navigation and planning. These datasets improve Otter's first-person scene interpretation and strategy formulation, especially for AR headset applications. The Otter model variant, tailored for egocentric views, marks a significant advancement in AR headset visual language modeling.

\begin{figure}[tp]
    \centering
    \begin{subfigure}[b]{\columnwidth}
        \centering
        \includegraphics[width=\textwidth]{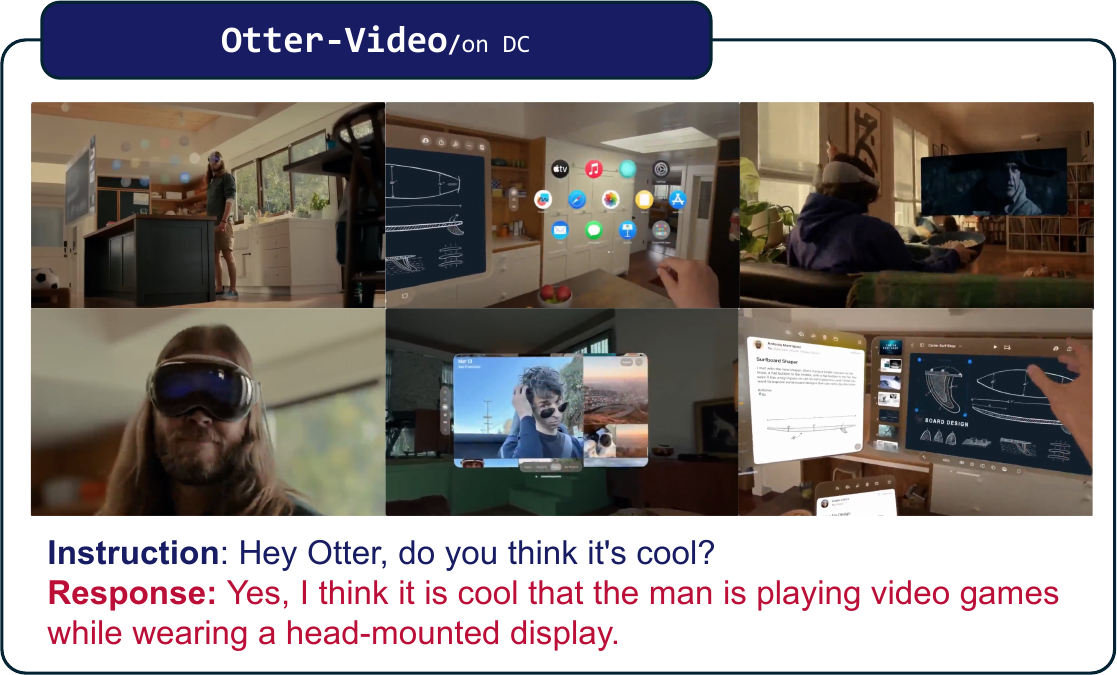}
        \label{fig:dc_demo}
    \end{subfigure}
    
    \vspace{1pt} 

    \begin{subfigure}[b]{\columnwidth}
        \centering
        \includegraphics[width=\textwidth]{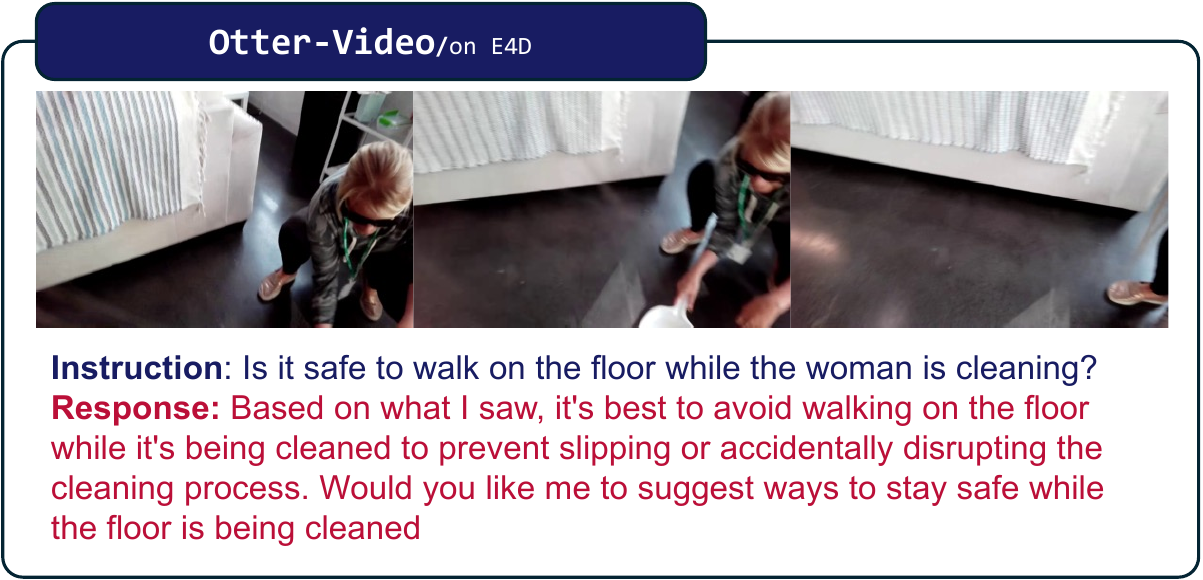}
        \label{fig:e4d_demo}
    \end{subfigure}
    \caption{Demonstrations of Otter's proficiency in different scenarios}
    \label{fig:otter_application}
\end{figure}

In~\cref{fig:otter_application}, we showcase Otter's adaptability in both scenarios, demonstrating its capabilities of responding to various instructions.

\subsection{Further Analysis}
\label{app:extended_exps}

\noindent \textbf{Task Descriptions}
\textcolor{black}{
For different subsets of MIMIC-IT, we generated task-specific descriptions with the assistance of ChatGPT. These descriptions serve as prefixes placed before the instruction/response pairs for each task, and brings potential benefits during the mixed training process of large data mixture to assist the model in understanding the required capabilities and response formats for different tasks. In this section, we provide the detailed information in generating task descriptions for each subsets.}

\textcolor{black}{For each dataset, we select 20 to 50 data entries (totaling less than 2048 tokens) for ChatGPT\footnote{Unless otherwise specified, the term \textit{ChatGPT} throughout our paper refers to \texttt{GPT-4-0613} model.} to summarize. The summarization by ChatGPT is requested to focus on three dimensions: (1) Coverage, (2) Ability, and (3) Answer Format. The query prompt used and the generated task description on MIMIC-IT's VST subset are as in~\cref{tab:task_description}
}


\noindent \textbf{Hyperparameters}
\textcolor{black}{We provide hyper-parameters related information at~\cref{tab:params}. The Otter models for Image and Video fundamentally employ similar hyperparameters. However, due to the need to accommodate multiple images (frames) in the Video model, it is necessary to reduce the corresponding batch size. We didn't conduct hyper-parameters search, the default configuration is sufficient to train good enough models for both image and video scenarios.
}

\begin{table}[t]
\centering
\caption{Summary of hyper-parameters of Otter-Image/Video.}
\label{tab:params}
\tabstyle{6pt}
\renewcommand{\arraystretch}{1.8}
\resizebox{0.7\columnwidth}{!}{%
\begin{tabular}{c|cc}
\textbf{Hyper-parameters} & \textbf{Otter-Image} & \textbf{Otter-Video} \\ \shline
Batch Size/GPU & 8 & 4 \\ \hline
LR & \multicolumn{2}{c}{1e-5} \\ \hline
LR Schedule & \multicolumn{2}{c}{cosine decay} \\ \hline
LR Warmup Ratio & \multicolumn{2}{c}{0.03} \\ \hline
Epoch & \multicolumn{2}{c}{6} \\ \hline
Sample Frames & - & 8 \\ \hline
Optimizer & \multicolumn{2}{c}{AdamW} \\ \hline
DeepSpeed & \multicolumn{2}{c}{Zero2} \\ \shline
Peak GPU Mem. & $\sim$70G & $\sim$76G
\end{tabular}
}
\end{table}



\begin{table}[t]
\centering
\caption{Ablations on adding data from other sources.}
\label{tab:other_sources}
\tabstyle{6pt}
\renewcommand{\arraystretch}{1.8}
\resizebox{0.7\columnwidth}{!}{%
\begin{tabular}{c|cc}
\textbf{Data Sources} & \textbf{MMBench} & \textbf{MM-Vet} \\ \shline
MIMIC-IT Subsets$^{*}$ & 41.5 & 19.7 \\ \hline
w/LLAVAR & 43.2 & 23.2 \\ \shline
w/Academic Datasets & \textbf{62.1} & \textbf{30.6}
\end{tabular}
}
\end{table}

\noindent \textbf{Video Evaluations}
\label{app:video_eval}
\textcolor{black}{In main paper, we provide the evaluation results on multiple Video-QA datasets. Here we elaborate on how we do the evaluations with ChatGPT as evaluator, as well as present results related to ChatGPT evaluation scores and other traditional metrics on Video-Caption tasks.}




\begin{table}[t]
\centering
\begin{lstlisting}[style=qaformat]
Q: what is the gender of the athlete?
A: the athlete is a male.
\end{lstlisting}
\end{table}

\textcolor{black}{Directly calculating matching accuracy may involve misconceptions during evaluations and can not reflect models true abilities to provide correct answers the questions. Consequently, many recent benchmarks~\cite{xie2023funqa,liu2023mmbench,yu2023mm} proposed to use ChatGPT as an evaluator to score by comparing the model's responses with the ground truth, and we have adopted this method as well in our evaluations for QA and Captioning tasks. In the following example, we demonstrate the query prompt used in the ChatGPT Evaluation process, as well as samples where ChatGPT scores are respectively 100 and 0. In the case of the 100-point model response, it precisely captures the information required in the Ground Truth (GT). In contrast, the 0-point response exhibits hallucination, resulting in a failure to score.}
\begin{table}[t]
\begin{lstlisting}[style=qaformat]
Prompt: You are a helpful AI assistant. Your task is to score the model response on a scale of 0 to 100 based on how well it matches the ground truth answer to the question.
\end{lstlisting}
\noindent\makebox[\columnwidth]{\rule{\columnwidth}{1pt}}
\begin{lstlisting}[style=qaformat]
Question: what is the gender of the athlete?
GT: the athlete is a male.
Response: The athlete shown in the video is male.
Score: 100
\end{lstlisting}
\noindent\makebox[\columnwidth]{\rule{\columnwidth}{1pt}}
\begin{lstlisting}[style=qaformat]
Question: what happened after the billiards?
GT: they chatted after playing billiards.
Response: After the man finishes explaining the game, the video shows clips of a man hitting the balls into the pockets on the table.
Score: 0
\end{lstlisting}
\end{table}

\noindent \textbf{Caption Results}
\textcolor{black}{Subsequently, in~\cref{tab:caption_results}, we provide results with both traditional metrics (BLEU, METEOR and ROUGE\_L) and ChatGPT scores for audience reference.} 

\begin{table}[htp]
\caption{Comparative results for captioning tasks on MSVD and MSRVTT with both traditional metrics and ChatGPT score. B-1: BLEU-1, M: METEOR, R: ROUGE\_L, G: ChatGPT Score}
\label{tab:caption_results}
\tabstyle{4pt}
\renewcommand{\arraystretch}{1.4}
\resizebox{0.95\columnwidth}{!}{%
\begin{tabular}{c|cccc}
\multirow{2}{*}{\textbf{Model}} & \multicolumn{2}{c}{\textbf{MSVD-Cap}} & \multicolumn{2}{c}{\textbf{MSVTT-Cap}} \\
 & \textbf{B-1/M/R} & \textbf{G} & \textbf{B-1/M/R} & \textbf{G} \\ \shline
Video-ChatGPT & 12.9/\textbf{15.8}/\textbf{21.4} & 46.4 & 10.5/20.5/33.7 & 33.1 \\ \hline
mPLUG-Owl-V & 13.0/14.2/19.5 & 51.5 & 10.5/23.9/41.5 & 39.0 \\ \shline
Otter-Video & \textbf{33.1}/14.3/21.0 & \textbf{54.3} & \textbf{58.3/32.6/59.0} & \textbf{67.9} \\
\end{tabular}
}
\end{table}

\begin{table}[htp]
\centering
\caption{Task description example of MIMIC-IT's VST subset.}
\label{tab:task_description}
\begin{lstlisting}[style=qaformat]
Prompt: I am writing a task description for a dataset. Can you take a look at few data from this task and provide me with the task description? The task description is used to describe the data to a vision language model. You may give a general summarization from the three perspectives (1) what is the questions and answers coverage of this task? (2) what ability needed to provide answers of given instructions in this task? (3) what is the general answer format of this task. Remember you can only observe a few items of this dataset so do not assume the data you observe is the whole, and keep your task description be general and summarized. You will be given multiple instances from the dataset in question/answer format. Please generate two versions of task description, one is short (less than 100 words) and one is long (less than 500 words).
\end{lstlisting}
\noindent\makebox[\columnwidth]{\rule{\columnwidth}{1pt}}
\begin{lstlisting}[style=qaformat]
Question: What color are the sheep?
Answer: The sheep are white.
Question: What are some of the toppings visible on the pizza?
Answer: Some of the toppings visible on the pizza include shrimp, basil, and possibly tomato sauce.
...
\end{lstlisting}
\noindent\makebox[\columnwidth]{\rule{\columnwidth}{1pt}}
\begin{lstlisting}[style=qaformat]
Short: This task involves answering questions based on detailed descriptions, observations, experiences, or events that the participants in the dataset have encountered. The questions cover a wide range of topics, from personal impressions of places and events to specific activities and historical significance. Answers require understanding of context and ability to provide informative responses based on the described scenarios. Answers are in several formats, including direct responses, narrative experiences, and explanatory comments.
Long: This task covers questions related to various experiences, observations, or events as relayed by individuals or inferred from provided details.
To address each data item effectively, you need a thorough comprehension of nuanced context and the ability to summarize experiences or observations. Answers might need to encapsulate personal reflections, summarize events, or describe the purpose and significance of activities. 
The answer format varies: some are narrative, detailing personal experiences or encounters; some are descriptive, painting a vivid picture of a scene or an event; and some are explanatory, providing reasons or background information about a subject.
You should tailor each answer to the specifics of the question, ensuring it is informative and accurately reflects the gist of the provided data, whether it conveys a subjective opinion, outlines a factual account, describes an experience, or explains the purpose of an event or organization.
\end{lstlisting}
\end{table}

\section{Conclusion}
\vspace{-1mm}
We presented the \textbf{Otter} model, a multi-modal model with in-context instruction tuning, driven by the comprehensive \textbf{MIMIC-IT} dataset with multi-modal contextual information and complex instruction-response pairs across diverse scenarios. Empirical results demonstrate Otter's superior training efficiency and performance in multimodal tasks. 

Otter has significant potentials to serve as a general-purpose assistant, yet it still has limitations in certain scenarios like fine-grained OCR tasks due to the relatively low-res (224$\times$224) of the pretrained vision encoder. Future research~\cite{fuyu-8b,li2023otterhd} on this, particularly in enabling models to more accurately perceive information from visual inputs, is highly anticipated.

\section*{Acknowledgement}
This study is supported by the Ministry of Education, Singapore, under its MOE AcRF Tier 2 (MOE-T2EP20221-0012, MOE-T2EP20223-0002), and under the RIE2020 Industry Alignment Fund – Industry Collaboration Projects (IAF-ICP) Funding Initiative, as well as cash and in-kind contribution from the industry partner(s).

\clearpage
{
\bibliographystyle{plain}
\bibliography{submission}
}

%

\begin{IEEEbiography}[{\includegraphics[width=1in,height=1.25in,clip,keepaspectratio]{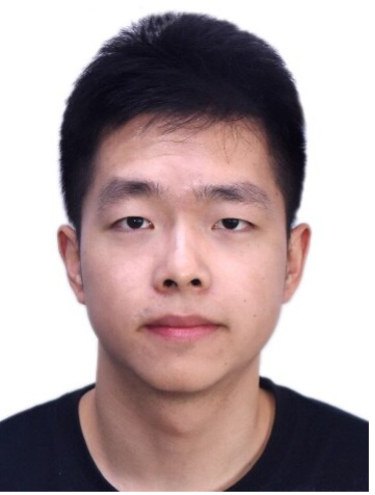}}]{Bo Li} received the B.S. from Harbin Institute of Technology in 2020. He is currently working towards the PhD degree in the College of Computer and Data Science, Nanyang Technological University, Singapore. His research interests mainly include multimodal learning and foundation models..
\end{IEEEbiography}

\begin{IEEEbiography}[{\includegraphics[width=1in,height=1.25in,clip,keepaspectratio]{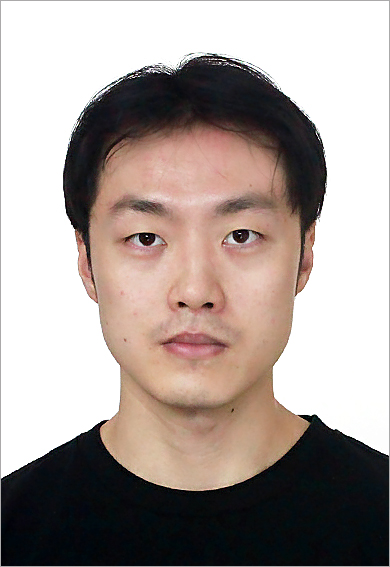}}]{Yuanhan Zhang} is currently a Ph.D. student at MMLab@NTU, Nanyang Technological University, supervised by Prof. Ziwei Liu. His interests lie in computer vision and deep learning. In particular, He is focused on adapting foundation models---from vision to multi-modal---for real-world exploration. He has published several papers in ICCV, ECCV, CVPR, NeurIPS and {\em IEEE
 Transactions on Pattern Analysis and Machine Intelligence} (TPAMI). He also served as a reviewer for CVPR, ICCV, ECCV, NeurIPS, ICML, ICLR, {\em IEEE
 Transactions on Pattern Analysis and Machine Intelligence} (TPAMI), and {\em International Journal of Computer Vision} (IJCV).
\end{IEEEbiography}

\begin{IEEEbiography}[{\includegraphics[width=1in,height=1.25in,clip,keepaspectratio]{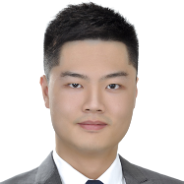}}]{Liangyu Chen} received his B.Eng. from Nanyang Technological University, Singapore, in 2022. He worked at MMLab@NTU from 2022 to 2024. He is currently pursuing a Ph.D. in Computer Science at Stanford University. He researches multimodal foundation models and data-centric machine learning.
\end{IEEEbiography}

\begin{IEEEbiography}[{\includegraphics[width=1in,height=1.25in,clip,keepaspectratio]{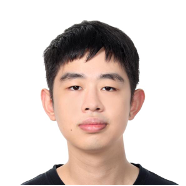}}]{Fanyi Pu} is an undergraduate student at Nanyang Technological University, Singapore, majoring in Data Science and Artificial Intelligence. His research focus is on multimodal large models.
\end{IEEEbiography}

\begin{IEEEbiography}[{\includegraphics[width=1in,height=1.25in,clip,keepaspectratio]{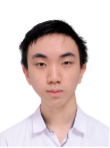}}]{Joshua Adrian} is an undergraduate student at Nanyang Technological University of Singapore, majoring in Data Science and Artificial Intelligence. His research focuses on multimodal models and agent based models.
\end{IEEEbiography}

\begin{IEEEbiography}[{\includegraphics[width=1in,height=1.25in,clip,keepaspectratio]{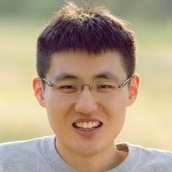}}]{Jingkang Yang} is a final-year PhD student at Nanyang Technological University (NTU), is working under the guidance of Professor Ziwei Liu. His research specializes in multimodal models and egocentric video understanding. Jingkang has published papers in top conferences such as CVPR, ICCV, ECCV, ICLR, and NeurIPS. Additionally, he has served as an outstanding reviewer for several top conferences, including CVPR, ICCV, ECCV, and as an area chair for ACL, EMNLP, and NAACL.
\end{IEEEbiography}

\begin{IEEEbiography}[{\includegraphics[width=1in,height=1.25in,clip,keepaspectratio]{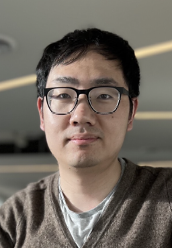}}]{Chunyuan Li}’s recent focus is large multimodal models in vision-and-language. His contributions include the development of LLaVA and the series of model families, as well as earlier works include Oscar, GLIP, Grounding DINO, GLIGEN and Florence. He has worked with xAI, ByteDance, Microsoft Research, and obtained his PhD at Duke University.
\end{IEEEbiography}

\begin{IEEEbiography}[{\includegraphics[width=1in,height=1.25in,clip,keepaspectratio]{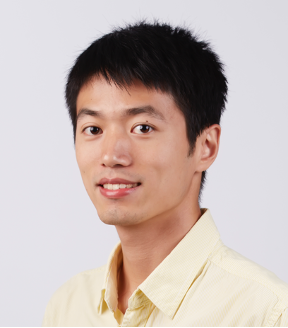}}]{Ziwei Liu} is currently an Associate Professor at Nanyang Technological University, Singapore. His research revolves around computer vision, machine learning and computer graphics. He has published extensively on top-tier conferences and journals in relevant fields, including CVPR, ICCV, ECCV, NeurIPS, ICLR, SIGGRAPH, TPAMI, TOG and Nature - Machine Intelligence. He is the recipient of PAMI Mark Everingham Prize, CVPR Best Paper Award Candidate, Asian Young Scientist Fellowship, International Congress of Basic Science Frontiers of Science Award and MIT Technology Review Innovators under 35 Asia Pacific. He serves as an Area Chair of CVPR, ICCV, ECCV, NeurIPS and ICLR, as well as an Associate Editor of IJCV.
\end{IEEEbiography}

\end{document}